\documentclass{ecai} 

\usepackage{latexsym}
\usepackage{amssymb}
\usepackage{amsmath}
\usepackage{amsthm}
\usepackage{booktabs}
\usepackage{enumitem}
\usepackage{graphicx}
\usepackage{color}
\usepackage{bm}
\usepackage{multirow}
\usepackage{enumitem}
\usepackage{float}
\usepackage{arydshln}
\usepackage{svg}
\usepackage{titletoc}
\usepackage{multicol}
\usepackage{makecell}

\newtheorem{theorem}{Theorem}

\newtheorem{proposition}[theorem]{Proposition}

\usepackage{array}
\newcolumntype{L}[1]{>{\raggedright\let\newline\\\arraybackslash\hspace{0pt}}m{#1}}
\newcolumntype{C}[1]{>{\centering\let\newline\\\arraybackslash\hspace{0pt}}m{#1}}
\newcolumntype{R}[1]{>{\raggedleft\let\newline\\\arraybackslash\hspace{0pt}}m{#1}}

\newcommand{\BibTeX}{B\kern-.05em{\sc i\kern-.025em b}\kern-.08em\TeX}

\definecolor{arsenic}{rgb}{0.23, 0.27, 0.29}

\usepackage{caption}
\begin{document}

\begin{frontmatter}

\paperid{835} 

\title{On the Improvement of Generalization and Stability of Forward-Only Learning via Neural Polarization}

\author[A]{\fnms{Erik B.}~\snm{Terres-Escudero}\thanks{Corresponding Author. Email: e.terres@deusto.es.}}
\author[B,C]{\fnms{Javier}~\snm{Del Ser}}
\author[A]{\fnms{Pablo}~\snm{Garcia-Bringas}} 

\address[A]{University of Deusto, 48007 Bilbao, Spain}
\address[B]{TECNALIA, Basque Research \& Technology Alliance (BRTA), 48160 Derio, Spain}
\address[C]{University of the Basque Country (UPV/EHU), 48013 Bilbao, Spain}

\begin{abstract}
Forward-only learning algorithms have recently gained attention as alternatives to gradient backpropagation, replacing the backward step of this latter solver with an additional contrastive forward pass. Among these approaches, the so-called Forward-Forward Algorithm (FFA) has been shown to achieve competitive levels of performance in terms of generalization and complexity. Networks trained using FFA learn to contrastively maximize a layer-wise defined goodness score when presented with real data (denoted as positive samples) and to minimize it when processing synthetic data (corr. negative samples). However, this algorithm still faces weaknesses that negatively affect the model accuracy and training stability, primarily due to a gradient imbalance between positive and negative samples. To overcome this issue, in this work we propose a novel implementation of the FFA algorithm, denoted as Polar-FFA, which extends the original formulation by introducing a neural division (\emph{polarization}) between positive and negative instances. Neurons in each of these groups aim to maximize their goodness when presented with their respective data type, thereby creating a symmetric gradient behavior. To empirically gauge the improved learning capabilities of our proposed Polar-FFA, we perform several systematic experiments using different activation and goodness functions over image classification datasets. Our results demonstrate that Polar-FFA outperforms FFA in terms of accuracy and convergence speed. Furthermore, its lower reliance on hyperparameters reduces the need for hyperparameter tuning to guarantee optimal generalization capabilities, thereby allowing for a broader range of neural network configurations.
\end{abstract}

\end{frontmatter}

\section{Introduction}

Biologically plausible algorithms are emerging as alternative learning approaches focused on addressing several well-known shortcomings inherent in the backpropagation algorithm (BP) \cite{zador2023catalyzing}. Among them, forward-only learning techniques stand out in the recent literature by leveraging error-driven local learning, thereby solving the weight transport and update lock problems \cite{ororbia2023brain}. These algorithms replace the backward pass of BP with an additional contrastive forward pass, modulated by carefully crafted layer-specific loss functions. Due to their local design, these algorithms allow training neural networks with a reduced memory footprint, suitable for scenarios with non-centralized computing capabilities, such as edge computing \cite{oguz2023forward,de2023mu}.

One of the most prominent algorithms within forward-only learning approaches is the so-called Forward-Forward Algorithm (FFA) \cite{hinton2022forward}. FFA advocates for the concept of fitness to contrastively learn to discriminate between real (also referred to as \emph{positive}) data and synthetic (corr. \emph{negative}) data. In doing so, FFA aims to maximize a goodness score when the network processes positive data, while minimizing this score when predicting negative data. Several works published since its inception have shown that FFA performs competitively when compared to BP. Unfortunately, FFA still faces several downsides that hinder the ability of this algorithm to achieve optimal generalization bounds. The cause of this weakness is primarily attributed to the formulation of the probability function that determines how the fitness score modulates whether a sample belongs to the positive set, as it has been shown to showcase vanishing gradient behavior \cite{lee2023symba}.

This work aims to advance towards addressing this issue by introduces Polar-FFA, a novel forward-only learning algorithm that extends the original FFA by incorporating neural polarization within each layer. This mechanism introduces the concept of positive and negative \emph{neurons}, which are shown to enhance the expressiveness of the probability function mentioned previously. We assess the benefits of Polar-FFA through extensive experiments over image classification datasets, showing that our approach enhances both the generalization capabilities of the trained network and the convergence speed of the learning process. Additionally, Polar-FFA is proven to allow for a broader set of neural configurations, thereby increasing the flexibility to build neural architectures based on FFA-like algorithms, which is critical when using bounded activation functions.

The rest of the manuscript is structured as follows: Section \ref{sec:2_related_work} introduces relevant literature to place in context the contribution of this work. Next, Section \ref{sec:3_characterization} motivates and describes the proposed Polar-FFA, together with examples of alternative probability functions. Section \ref{sec:4_exp_setu} follows by posing research questions and the experimental setup used to inform their responses with evidence. Section \ref{sec:5_results} presents the obtained experimental results and discusses on the improvements and limitations of our method. Finally, Section \ref{sec:6_concl} draws the main conclusions of the work and outlines potential research directions rooted in our findings here reported.

\section{Related Work}
\label{sec:2_related_work}

Before proceeding with the description of Polar-FFA, we briefly overview prior work on forward-only learning and FFA, ending with a statement of the contribution of Polar-FFA to the state of the art:

\paragraph{Forward-only Learning} BP is arguably the most widely used algorithm for training neural networks. However, a recent surge in neuro-inspired learning algorithms has gained momentum in the Artificial Intelligence community \cite{zador2023catalyzing}. These algorithms aim at addressing well-known algorithmic weaknesses of other learners by studying the learning dynamics in biological brains, including the usage of sparse latent activity and local learning rules. As a result, neuro-inspired learning algorithms have achieved competitive generalization capabilities \cite{illing2021local,moraitis2022softhebb}. Among them, forward-only learning techniques present a novel credit assignment mechanism heavily inspired by the learning dynamics present in Hebbian update rules. They replace the backward pass of BP with a secondary forward pass, used in a layer-wise manner, to contrastively learn relevant features from input data \cite{ororbia2023brain}. The first implementation of this technique can be attributed to Kohan et al. \cite{kohan2018error}, whose approach involved connecting the obtained classification error with the input layer. This allows the network to forward this error during a second forward pass, updating the weights without employing backward connections. An alternative forward-only approach was developed by Dellaferrera \& Kreiman in \cite{dellaferrera2022error}, where a novel error-driven update rule was proposed to modulate the input perturbation and contrastively train each layer. 

\paragraph{Forward-Forward Algorithm} Further within the family of forward-only learning algorithms, FFA is a recently proposed neuroinspired approach based on the maximization of a \emph{layer fitness} \cite{hinton2022forward}. In doing so, FFA resorts to a contrastive learning process, where models are trained to distinguish between real (\emph{positive}) data and synthetic (\emph{negative}) data. To this end, FFA requires the definition of i) a \emph{goodness function}, which measures the fitness of a sample to belong to the positive set of data; and ii) a probability function, which is used to map the fitness scores to the range $\mathbb{R}[0,1]$. Formally, a goodness function $G: \mathbb{R}^n \mapsto \mathbb{R}[0,\infty)$ maps a latent vector $\bm{\ell} \in \mathbb{R}^n$ to a non-negative fitness value. Common choices for the goodness function in the literature include the square Euclidean norm:
\begin{equation}
    G(\bm{\ell}) = \|\bm{\ell}\|^2_2 = \sum_{i=1}^n \ell_i^2,
\end{equation}
where $\bm{\ell}=(\ell_1,\ldots,\ell_n)$. Building upon this goodness function, FFA utilizes a probability function $P : \mathbb{R}[0,\infty]\mapsto \mathbb{R}[0,1]$, enabling the use of probabilistic loss functions (e.g. binary cross entropy). In his seminal work, Hinton suggested using a sigmoidal function as this mapping, with a hyper-parameter $\theta$ that shifts the center of the distribution: 
\begin{equation}
P(G(\bm{\ell})) = \sigma(G(\bm{\ell}); \theta) = \frac{1}{1+e^{-G(\bm{\ell})+\theta}}.
\end{equation}

Due to its layer-wise dynamics, FFA emerges as a highly competitive alternative to other learning algorithms, especially in scenarios where memory and energy are highly constrained. For example, this algorithm has found practical applications in two relevant edge systems: optical neural networks, achieving competitive accuracy with a reduced number of parameters \cite{oguz2023forward}; and microcontrollers, enabling on-device training for multivariate regression tasks \cite{de2023mu}. Some extensions of the algorithm have been proposed to incorporate greater biological plausibility, such as the integration of predictive coding heuristics \citep{ororbia2023predictive}, and its adaptation to spiking neural networks \citep{ororbia2023contrastive}. In their work on predictive FFA (PFFA) \cite{ororbia2023predictive}, Ororbia \& Mali also highlighted an additional key property of models trained with FFA: the resulting latent space is composed of distinct clusters consisting of points of the same class. A similar effect was exposed by Tosato et al. in \cite{tosato2023emergent}, underscoring the apparent sparsity of latent vectors and its inherent latent structure. This effect was further explored by Yang \cite{yang2023theory}, who provided a mathematical foundation for this phenomenon in ReLU-based networks using squared Euclidean distance as a goodness function.

\section{Proposed Polar Forward-Forward Algorithm}
\label{sec:3_characterization}

\begin{figure*}[]
    \centering
    \includegraphics[width=2\columnwidth]{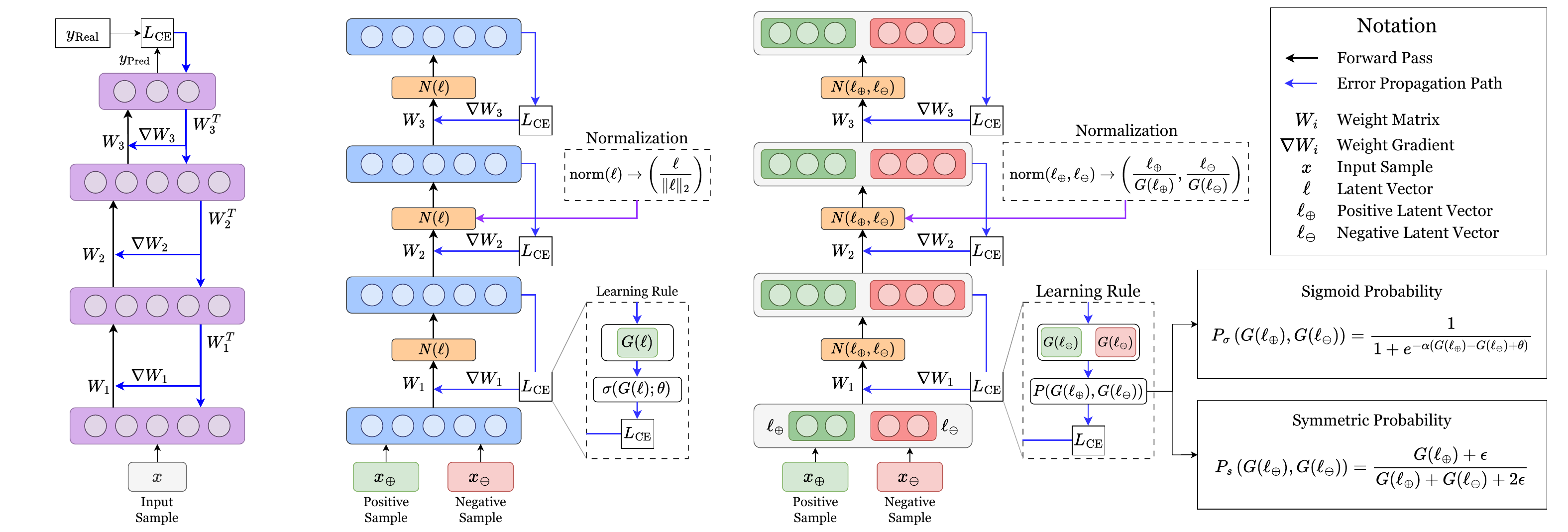}
    \resizebox{1.8\columnwidth}{!}{\begin{tabular}{C{0.2\columnwidth}C{0.5\columnwidth}C{1.15\columnwidth}}
         (a) BP. & (b) FFA. & (c) Proposed Polar-FFA. 
    \end{tabular}}
    \caption{Forward and backward propagation's paths on (a) Backpropagation (BP); (b) Forward-Forward Algorithm (FFA); and (c) Polar-FFA. Black lines denote the forward direction of the information flowing from the input through each of the networks. Blue lines indicate the error BP path, which has a local behavior in FFA and Polar-FFA. Additionally, the two adapted probability functions and the normalization process are included in the plot.}
    \label{fig:diferent_approaches}
    
\end{figure*}

Polar-FFA introduces an extension to the FFA formulation by integrating a neural division where each neuron is assigned either a positive or negative polarization. The fundamental learning mechanism remains quite similar to FFA, as neurons within each set are trained to maximize their goodness score when exposed to samples of their corresponding polarity, and to minimize it when presented with the opposite polarity. For example, when employing a activity based goodness function, a positive neuron is expected to maximize its activity when presented with positive data, and to minimize it when presented with negative samples. However, due to this neural partitioning, the probability function measuring whether sample belongs to the positive set must be adapted from a single goodness score to a formulation including positive and negative goodness values. To provide a formal description of our algorithm, we recall the theoretical framework of FFA outlined in Section \ref{sec:2_related_work} for the sake of consistent notation and conceptual clarity. Since FFA-like algorithms train models on a layer by layer basis, the formulation of the proposed Polar-FFA focuses on the mechanisms involved in training a single layer. We hereafter denote the set of neurons in a given network layer as $\mathcal{L}$, so that $\bm{\ell}$ refers to the latent vector at the output of the layer at hand.

% Division y goodness
% A priori bien
The definition of Polar-FFA departs from the  assignment of a polarity to each neuron, depending on the expected goodness behavior desired for this neuron. Similarly to the original FFA, we define the subset of positive neurons as $\mathcal{L}_{\oplus}$, which aims at maximizing its goodness score when exposed to positive samples. The novel concept introduced in Polar-FFA is the negative neural set, denoted as $\mathcal{L}_{\ominus}$, which, in contrast to its counterpart, aims to maximize its goodness score when presented with negative samples. While the relative sizes of $\mathcal{L}_{\oplus}$ and $\mathcal{L}_{\ominus}$ can be arbitrarily specified whenever $\mathcal{L}_{\oplus}\cup \mathcal{L}_{\ominus}=\mathcal{L}$, for the sake of simplicity in this paper, we limit our discussion to scenarios where $|\mathcal{L}_{\oplus}| = |\mathcal{L}_{\ominus}|$. Under this split architecture, the goodness score is reformulated from a single scalar measuring the fitness of the input within the positive data distribution, to a pair of goodness scores, each measuring the suitability of the input with respect to the data distribution of their respective polarity. Since the goodness function only processes information contained in the latent vector, Polar-FFA can naturally consider the same set of goodness functions as those considered for FFA in the literature. Consequently, the goodness function $G:\mathbb{R}^n\mapsto \mathbb{R}[0,\infty)\times \mathbb{R}[0,\infty)$ evaluates each group independently as:
\begin{equation}
    \label{eq:Polar_goodness_of_layer}
    G(\bm{\ell}) = G(\bm{\ell}_{\oplus}\cup \bm{\ell}_{\ominus}) = \left(G(\bm{\ell}_{\oplus}), G(\bm{\ell}_{\ominus})\right),
\end{equation}
where $\bm{\ell} = \bm{\ell}_{\oplus}\cup\bm{\ell}_{\ominus} \in \mathbb{R}^{n}$ is the latent vector at the output of layer $L$, which contains $n$ neurons, and $\bm{\ell}_{\oplus}$ ($\bm{\ell}_{\ominus}$) denote the activations corresponding to positive (negative) neurons in $\mathcal{L}_\oplus$ ($\mathcal{L}_\ominus$).

% Probability y loss
% A priori bien
The second step in the adaptation from FFA to Polar-FFA involves replacing the scalar-based probability function with a probability function $P : \mathbb{R}[0,\infty) \times \mathbb{R}[0,\infty)\mapsto \mathbb{R}[0, 1]$, which receives a pair of goodness scores at its input. This function should maximize its value as the discrepancy between positive and negative goodness scores increases. Once  this probability function is defined, Polar-FFA can be trained under the Binary Cross-Entropy (BCE) loss $L_{\textup{CE}}$ using an analogous training process as the standard FFA. Formally, the loss function of any forward-like supervised learning process is given by:
\begin{equation}
L_{\textup{CE}} = \mbox{--}\sum_{i} \rho_i\log\left[P(G(\bm{\ell}_i))\right] + (1-\rho_i)\log\left[1-P(G(\bm{\ell}_i))\right],
\end{equation}
where $\rho_i\in\{0,1\}$ is a binary variable referring to the polarity of the data, taking value $1$ if the sample is positive and $0$ otherwise; and $\bm{\ell}_i$ extends the above notation to refer to the latent vector obtained by processing the $i$-th input. Using this loss function, weight updates at each step of Polar-FFA can be manually computed by applying the chain rule. Given a weight $w_{ij}$ associated to a neuron that belongs to the positive neural set $\mathcal{L}_{\oplus}$, the loss gradient yields as:
\begin{equation}
    \label{eq:weight_update}
    \frac{\partial L_{\textup{CE}}}{\partial w_{ij}}= - \sum_{k} \frac{1}{P(G(\bm{\ell}_{\oplus}^k), G(\bm{\ell}_{\ominus}^k))}\frac{\partial P}{\partial G_{\oplus}} \frac{\partial G_{\oplus}}{\partial \ell_{\oplus,j}^k} \frac{\partial \ell_{\oplus,j}^k}{\partial a_{\oplus, j}^k} x_{i}^k,
\end{equation}
where we expand the notation of the latent vector further as $(\bm{\ell}_{\oplus}^k)_j=\ell_{\oplus,j}^k$ to denote the activation vector obtained from forwarding the input sample $\mathbf{x}^k$, $(\mathbf{a}_{\oplus}^k)_j=a_{{\oplus},j}^k$ refers to the preactivation latent vector, and we employ the $\square_j$ subindex to refer to the $j$-th coordinate of the respective vector. Similarly, we abreviate $G(\bm{\ell}^k_{\oplus})$ to $G_{\oplus}$. An equivalent expression of the loss can be obtained for the weight update rule of neurons in $\mathcal{L}_{\ominus}$ by replacing $\bm{\ell}_{\oplus}$ by $\bm{\ell}_{\ominus}$. It is important to note the heavy symmetry present in these expressions, as the behavior of neurons in the negative neural set with negative data corresponds to the same maximization objective as positive neurons within the positive dataset.

\paragraph{Characterization of FFA and Polar-FFA} As shown in Expression \eqref{eq:weight_update}, any update rule in a FFA-like algorithm is controlled by three key factors: the probability function, the goodness function, and the activation function. Therefore, understanding the behavior of any standard FFA-like model boils down to examining the specific behavior of these three functions. We refer as \textit{network configurations} to any combination of activation, goodness and probability function used to define and train a specific model. Within this configuration, various choices are possible; for instance, any norm can serve as a goodness function, whereas any within the plethora of neural activation functions defined in the literature can be employed at each layer. This categorization has been crucial in providing a comprehensive set of neural configurations for the experimental setup, as is further detailed in Appendix \ref{app:rq1_model_config_details}.

\paragraph{Layer normalization} Given the layer-wise learning dynamics of FFA and Polar-FFA, it is crucial to ensure that goodness information from previous layers is not overly influential in the decision making of subsequent layers. This issue was addressed in the original FFA work, where Hinton proposed introducing a normalization process between layers to equalize the goodness scores of all latent vectors \cite{hinton2022forward}. Since our paper explores a broader set of goodness functions, this approach must be expanded to guarantee that this property is preserved for any such choice. Since the probability function in Polar-FFA only analyzes the relationship between positive and negative goodness values, the normalization scheme must involve updating the positive and negative latent vectors to yield equal goodness scores. This transformation ensures that latent vectors are treated equally by the probability function, removing any bias in subsequent layers. To avoid additional complexity, we restrict goodness functions to be absolutely homogeneous, meaning that $G(\lambda\bm{\ell}) = |\lambda| G(\bm{\ell})$ $\forall \bm{\ell} \in \mathbb{R}^n$ and $\forall \lambda \in \mathbb{R}$. Subject to this constraint, we can verify that the following normalization function $\texttt{norm}:\mathbb{R}^n \mapsto \mathbb{R}^n$ satisfies the sought normalization properties:
\begin{equation}
    \texttt{norm}(\bm{\ell}_{\oplus}, \bm{\ell}_{\ominus}) = \left( \frac{\bm{\ell}_{\oplus}}{G(\bm{\ell}_{\oplus})}, \frac{\bm{\ell}_{\ominus}}{G(\bm{\ell}_{\ominus})}\right),
\end{equation}
namely, the latent vectors corresponding to positive and negative neurons are normalized by their associated goodness value.

\paragraph{Definition of probability function}

Due to the distinct formulation of the probability function in Polar-FFA, the standard sigmoidal probability function cannot be directly employed to train the model. As previously discussed, the probability function in Polar-FFA must yield a probability based on the relationship between the pair of goodness scores $(G(\bm{\ell}_{\oplus}), G(\bm{\ell}_{\ominus}))$ at its input. High probability scores (values close to 1) should be produced when presented with scores satisfying $G(\bm{\ell}_{\oplus}) \gg G(\bm{\ell}_{\ominus})$, and conversely, low probability values (close to 0) when $G(\bm{\ell}_{\oplus}) \ll G(\bm{\ell}_{\ominus})$.

We present two alternative probability functions that leverage different relationships between positive and negative goodness scores, thus creating two distinct learning dynamics. Our first proposal expands the original sigmoid probability to incorporate negative goodness. The second approach involves computing the ratio of positive goodness to the total goodness score:
\begin{enumerate}[leftmargin=*]
\item \textbf{Polar sigmoid probability} $P_\sigma(\cdot)$: The first proposed probability function extends the original sigmoidal probability in FFA by substituting the value $G(\bm{\ell})$ with the difference in scores $G(\bm{\ell}_{\oplus}) - G(\bm{\ell}_{\ominus})$, thereby determining the probability based on the disparity between positive and negative goodness values. We denote this probability function as:
\begin{equation}
P_\sigma\left(G(\bm{\ell}_{\oplus}), G(\bm{\ell}_{\ominus}); \alpha, \theta\right) = \frac{1}{1+e^{-\alpha\cdot (G(\bm{\ell}_{\oplus})-G(\bm{\ell}_{\ominus})+\theta)}}.
\end{equation}

Similarly to the original sigmoid probability in FFA, $\alpha$ scales the difference in activity, whereas $\theta$ shifts the center of the function. This function exhibits a learning behavior similar to its predecessor, but incorporates two crucial properties. Firstly, since the function is defined by the difference of two opposite activities, the probability function is not biased towards the positive probability, thereby avoiding the gradient asymmetry identified by Lee et al. \cite{lee2023symba}. Secondly, as the split version does not directly employ the latent activity, the model's learning is independent of the mean of the latent activity, increasing the flexibility of the architecture. The effect of the mean on the learning dynamics is thoroughly examined in Appendix \ref{app:prq_unbiased_sigmoid_probability}.

Notably, our formulation of $P_\sigma(\cdot)$ is solely influenced by the variance of the sigmoidal value of the difference between the positive and negative goodness values. Under these conditions, the mean value of the derivative can be guaranteed to be greater than any small number if the value of the difference in the goodness has low variance. This theoretical result implies that models with naturally low variance can learn, independently of their mean value. These insights are mathematically stated in Proposition \ref{prop:split_sigmoid_unbiased}, which is mathematically proven in Appendix \ref{app:proposition_1_proof}:

\begin{proposition}
    \label{prop:split_sigmoid_unbiased}
    Let $\mathbf{W} \in \mathbb{R}^{n \times m}$ be a randomly initialized weight matrix, containing two independent sub-matrices of weights $(\mathbf{W}_{\oplus}, \mathbf{W}_{\ominus})$ connecting an input $\mathbf{x}$ to their respective neural groups. Let $f(\cdot)$ be a continuous activation function, and let $\bm{\ell} = (\bm{\ell}_{\oplus}, \bm{\ell}_{\ominus}) = \left(f(\mathbf{W}_{\oplus}\mathbf{x}^T), f(\mathbf{W}_{\ominus}\mathbf{x}^T)\right) = f(\mathbf{W} \mathbf{x}^T)$ be the output vector of the layer. Let $z$ be a random variable defined by $z = G(\bm{\ell}_{\oplus})-G(\bm{\ell}_{\ominus})$, such that $P_\sigma(G(\bm{\ell}_{\oplus}),G(\bm{\ell}_{\ominus});\alpha,\theta)\equiv P_\sigma (z;\alpha,\theta)$. Then, for $\theta=0$ and $\alpha=1$:
    \begin{enumerate}[leftmargin=*]
        \item The expected value of the derivative sigmoid probability function is determined by the variance of the sigmoid of $z$ as:
        \begin{equation}
            \mathbb{E}\left[\ \frac{\partial P_\sigma(z)}{\partial G(\bm{\ell}_{\oplus})} \right] = \frac{1}{4} - \textup{Var}\left[P_\sigma(z)\right] \geq 0.
        \end{equation}
        \item The expected value of the derivative is bounded below by the variance of $z$:
        \begin{equation}
            \mathbb{E}\left[\ \frac{\partial P_\sigma(z)}{\partial G(\bm{\ell}_{\oplus})} \right] \geq \frac{1}{4}- \textup{Var}[z].
        \end{equation}
    \end{enumerate}
\end{proposition}

\item \textbf{Symmetric probability} $P_s(\cdot)$: The second proposed probability function replaces the difference in goodness with the ratio between the positive goodness and the sum of the two scores:
\begin{equation}\label{eq:symmetric_prob}
P_s(G(\bm{\ell}_{\oplus}), G(\bm{\ell}_{\ominus})) = \frac{G(\bm{\ell}_{\oplus}) + \epsilon}{G(\bm{\ell}_{\oplus})+ G(\bm{\ell}_{\ominus}) + 2\epsilon},
\end{equation}
where $\bm{\ell}$ follows the previously introduced notation for latent vectors, and $\epsilon$ is a small number introduced to avoid division by zero. The motivation behind adding $\epsilon$ in both the numerator and the denominator, with the denominator being doubled, is to ensure that the function remains symmetric for negative samples in latent vectors with low activity. This symmetry is guaranteed by the expression $P_s(G(\bm{\ell}_{\oplus}), G(\bm{\ell}_{\ominus})) = 1 - P_s(G(\bm{\ell}_{\ominus}), G(\bm{\ell}_{\oplus}))$.

Similar to the previously defined sigmoid probability function $P_\sigma(\cdot)$, this function maintains stable initial learning dynamics, ensuring that its gradient is nonzero during early training. Furthermore, a more robust theoretical result can be attributed to this function, as the update behavior is driven by the ratio between negative and positive activities, which is tightly related to the model's accuracy. For instance, the only points at which the derivative of this function is close to zero are when the model is remarkably accurate or when the values of the goodness score reach large values. The second situation can be mitigated by either clipping the goodness value or by including additional regularization terms into the loss function. Therefore, as opposed to models trained with FFA using the sigmoid function, models with low accuracy levels are ensured not to get stuck during the forward-only learning process  regardless of their activity or variance, thereby making this function theoretically more robust. This property is formally expressed in Proposition \ref{prop:symmetric_probability_never_zero}, proven in Appendix \ref{app:proposition_2_proof}.
\begin{proposition}
\label{prop:symmetric_probability_never_zero}
    Let $\bm{\ell} = (\bm{\ell}_{\oplus}, \bm{\ell}_{\ominus})$ be a latent vector, and let $G(\cdot)$ be an arbitrary goodness function. The following properties hold:
    \begin{enumerate}[leftmargin=*]
        \item $P_s(G(\bm{\ell}_{\oplus}), G(\bm{\ell}_{\ominus}))$ is approximately scale invariant for all $G(\bm{\ell}_{\oplus})$ such that $G(\bm{\ell}_{\ominus}) \gg \epsilon$.
        \item Given two values $a,b \in \mathbb{R}[0,\infty)$ such that $a < G(\bm{\ell}_{\oplus})+G(\bm{\ell}_{\ominus}) < b$, then the order of growth of the derivative is:
        \begin{equation}
           \hspace{-6mm}\frac{\partial P_s(G(\bm{\ell}_{\oplus}), G(\bm{\ell}_{\ominus}))}{\partial G(\bm{\ell}_{\oplus})} = \mathcal{O}\left(\frac{G(\bm{\ell}_{\ominus})}{G(\bm{\ell}_{\oplus})}\right),
        \end{equation}
        where $\mathcal{O}(\cdot)$ denotes asymptotic complexity.
    \end{enumerate}
\end{proposition}
\end{enumerate}

\section{Experimental Setup}
\label{sec:4_exp_setu}
To empirically assess the performance of our Polar-FFA approach, we formulate two Research Questions (RQs) that will be analyzed through an extensive set of experiments:
\begin{itemize}[leftmargin=*]
\item \textbf{RQ1}: \textit{Does neural polarization enhance the convergence and generalization with respect to the original FFA?}
\item \textbf{RQ2}: \textit{Which insights can be obtained by analyzing the latent space induced by neural polarization?}

\end{itemize}

To ensure that the results within RQ1 are not biased towards specific network configurations, exhaustive tests have been done with both FFA and Polar-FFA across a diverse range of architectural configurations. These configurations yield from the combinations of 3 different activation functions (\texttt{ReLU}, \texttt{Sigmoid}, and \texttt{Tanh}), $24$ goodness functions $G(\cdot)$ (including $\|\cdot \|_2$ and $\|\cdot\|_1$, among others) and $3$ probability functions, namely, the original sigmoid function used in FFA, hereafter denoted as $P_\sigma^{\textup{FFA}}(\cdot)$, and the proposed $P_\sigma(\cdot)$ and $P_s(\cdot)$. The detailed list of network configurations utilized for experiments related to RQ1 is reported in Appendix \ref{app:rq1_model_config_details}. 

The selected datasets for the experiments include MNIST \cite{lecun1998gradient}, Fashion MNIST \cite{xiao2017fashion}, KMNIST \cite{clanuwat2018deep}, and CIFAR-10 \cite{krizhevsky2009learning}. Models trained on MNIST-like datasets (MNIST, Fashion MNIST, or KMNIST) use a 2-layer architecture comprising 1000 neurons each, whereas models trained on CIFAR-10 contain 2000 neurons per layer. In networks trained via Polar-FFA, the first half of the neurons of each layer are assigned to the positive neural group $\mathcal{L}_\oplus$ and the second half to the negative set $\mathcal{L}_\ominus$, as this polarity distribution can be empirically shown to be the most stable (see Appendix \ref{app:ratio_neurons}). All models are trained using the ADAM optimizer with a learning rate of $0.001$ for $50$ epochs and a batch size of $512$, except for CIFAR-10, in which the training process is extended to $100$ epochs. Following \cite{lee2023symba}, labels are embedded into images by concatenating them to the end of the image, using a Bernoulli distribution with probability $0.1$ and a label pattern size of $100$ pixels. Furthermore, we adopt Hinton's original approach to generate negative data instances \cite{hinton2022forward}, which involved embedding incorrect data (selected from the set of possible classes) to a real positive instance. Each (activation,goodness,probability) function combination is used to train the network on each dataset using FFA and Polar-FFA.

Each of these configurations is evaluated for accuracy and convergence speed. As for the latter, we rely on what we denote as \emph{convergence area} ($\textup{CA}$), defined as the area above the accuracy curve across epochs, upper bounded by the maximum accuracy attained over all epochs. Formally, given the accuracy levels $\mathbf{acc} = (\textup{acc}_1, \dots, \textup{acc}_T)$ obtained by the learning algorithm at hand over $T$ epochs, this score is given by:\begin{equation}
    \textup{CA}(\mathbf{acc}) = \frac{1}{T\cdot \max_t \textup{acc}_t} \sum_{i=t}^T \left[(\max\nolimits_t \textup{acc}_t) - \textup{acc}_t \right].
\end{equation}

% A priori bien
To evaluate the statistical significance of the performance gaps between FFA and Polar-FFA, we select a diverse set of neural configurations and trained them using $10$ different seeds over the MNIST, KMNIST, and Fashion MNIST datasets. Accuracy measurements are conducted for all three experiments to calculate their generalization capability and its standard error of the mean. The selection of these configurations is independently performed for each probability function. Specifically, we select the two goodness functions with the highest average accuracy on MNIST-like datasets for each of the three activation functions.

% REvisar
RQ2 aims to offer deeper insights and arguments regarding the results obtained in response to RQ1. This analysis is conducted by exploring the latent space of models trained using the previously defined set of neural configurations. By extracting a representative sample of their latent space, we compute several geometric properties, primarily focusing on sparsity and separability indices. These indices are then inspected jointly with the accuracy and convergence speed of the trained network to analyze the relationship between the geometric properties of the latent space and the training dynamics of the networks. Moreover, we resort to T-SNE \cite{van2008visualizing} as a dimensionality reduction algorithm to visualize the distribution of the latent space $\bm{\ell}$.

% A priori bien
The sparsity of the latent space is quantified in terms of the Hoyer metric \cite{hoyer2004non}, which is known to meet most properties expected for sparsity metrics \cite{hurley2009comparing}. Given a latent vector $\bm{\ell} \in \mathbb{R}^{n}$, this metric returns a value between $0$ and $1$ measuring the sparsity of the vector, where $0$ implies a uniform distribution and $1$ a totally sparse vector. The Hoyer Index $\textup{HI} : \mathbb{R}^{n} \rightarrow \mathbb{R}[0,1]$ is defined as:
\begin{equation}
\textup{HI}(\bm{\ell}) = \left(\sqrt{n} - (\|\bm{\ell}\|_1/\|\bm{\ell}\|_2)\right)/\left(\sqrt{n}-1\right).
\end{equation}

Based on this index, we introduce a score denoted as \emph{neural usage}, which measures how well distributed is the contribution of the different neurons to the latent output of the layer. We define it as the Hoyer index of the latent vector obtained by averaging over the output latent space. A highly sparse average latent vector would imply the existence of a large set of neurons that do not actively participate  in the inference forward pass of the model, while low sparsity values would indicate that all neurons contribute equally to the prediction of the model. It is important to note that this score is particularly relevant for FFA-like methods, where the activity of neurons directly influences model predictions.

% A priori bien
Similarly, separability focuses on the distance in the latent space between clusters of different classes. In FFA, this score quantifies the overlap between positive and negative latent vectors. To achieve this, we use the geometric separability index \cite{thornton1998separability}. This metric analyzes the ratio of samples belonging to the same class as a given point within their nearest neighbors. Given a dataset composed of pairs of samples and labels, we denote the class of the $k$-th sample as $C_k$. Similarly, we denote the polarity type of its $j$-th nearest neighbor as $\textup{KNN}_j(C_k)$. Values of separability closer to $1$ imply that the different classes do not overlap, while values closer to $0$ imply a total overlap of the classes. The separability index $\textup{SI}(\bm{\ell})$ is computed as:
\begin{equation}
\textup{SI}(\bm{\ell}) = \frac{1}{K\cdot J}\sum_{k=1}^{K} \sum_{j=1}^J \delta(C_k, \textup{KNN}_j(C_k)),
\end{equation}
where $K$ represents the number of input samples, and $J$ is the total number of neighbors selected, which we set to $J=5$ for our experiments. Additionally, $\delta(x,y)$ denotes the Kronecker delta function, which outputs $1$ if $x=y$ and $0$ otherwise. 

All the source code for the project can be accessed via \url{https://github.com/erikberter/PolarFFA}.

\section{Results and Discussion}
\label{sec:5_results}

In this section, we present and discuss on the results obtained for each of the previously introduced research question:

\subsection*{RQ1: Does neural polarization enhance the convergence and generalization with respect to FFA?}
\label{sec:subsec_rq1}
%% Metrics on accuracy
The results of the average accuracy of the distinct models on MNIST, Fashion MNIST and KMNIST are presented in Table \ref{tab:train_overall_results}. Due to the large accuracy difference between these datasets and CIFAR-10, the results corresponding to the latter are presented in Table \ref{tab:train_overall_results_cifar10}. However, the disaggregated results of all the experiments can be found in Table \ref{tab:total_accuracies} in Appendix \ref{app:rq1_ablation_study}.
\begin{table}[h]
    \centering
    \caption{Maximum, median, average and minimum values of the accuracy [\%] of the different probability functions of FFA and Polar-FFA, averaged over the MNIST, Fashion-MNIST and KMNIST datasets. We also present the amount of configurations achieving an accuracy greater than $80\%$ and the total number of configurations. Best results for each score in bold.}
    \label{tab:train_overall_results}
    
    \resizebox{\columnwidth}{!}{\begin{tabular}{lccc}
\toprule[1pt]\midrule[0.3pt]
Score     & $P_{s}$ & $P_{\sigma}$     & $P_\sigma^{\textup{FFA}}$     \\
\midrule
 Maximum accuracy                        &       91.60  &   \textbf{92.89} &   90.76 \\
 Median accuracy                      &       \textbf{89.67} &   86.29 &   74.76 \\
 Average accuracy                    &       \textbf{86.09} &   77.68 &   58.18 \\
 Minimum accuracy                        &       \textbf{68.90}  &   40.43 &    7.64 \\
 \midrule
\# top-80\% accuracy configurations/total & \textbf{30/36} & 20/36 & 13/36   \\
Convergence area top-80\% configurations  &  \textbf{0.0122} & 0.0177 & 0.0208\\
\midrule[0.3pt]\bottomrule[1pt]
\end{tabular}}   
\end{table}

% A priori bien | x2
The results obtained for this first research question confirm our hypothesis regarding the improved performance of Polar-FFA compared to FFA. Firstly, analyzing the accuracy scores obtained for MNIST-like datasets in Table \ref{tab:train_overall_results}, it is evident that models trained using Polar-FFA outperform those trained using FFA in terms of accuracy, especially in cases where the neural configuration renders the model incapable of learning, as detailed in Appendix \ref{app:prq_unbiased_sigmoid_probability}. Moreover, when comparing the results between $P_{\sigma}$ and $P_{\sigma}^{\textup{FFA}}$ individually, a significant number of models outperform their FFA counterparts. In terms of robustness, Polar-FFA demonstrates a significant advantage over FFA, with more than a 30-point difference in average accuracy observed in the worst-performing network configuration. This difference showcases the importance of ensuring that the derivative of the probability achieves non-zero values, as highlighted in Proposition \ref{prop:split_sigmoid_unbiased} and Proposition \ref{prop:symmetric_probability_never_zero}. Notably, while the highest generalization capabilities are attributed to $P_\sigma$, the symmetric probability function $P_s$ achieves the most robust results, with a remarkable minimum average accuracy of $68.90$\% in MNIST-like datasets, surpassing both sigmoidal functions by more than 28 points of accuracy. Furthermore, this minimal accuracy value increases to $83.92$\% when using a lateral inhibition scheme, demonstrating high accuracy across all configurations. This trend is also reflected on the total number of models reaching the minimum accuracy of $80$\%, where FFA models are revealed to attain poor generalization capabilities. This effect highlights that the scale invariance of the probability function significantly alleviates the activity constraints required for sigmoid-like probability functions. Similarly, the theoretical effect stated in Proposition \ref{prop:split_sigmoid_unbiased} is also observed in these results, as the minimum accuracy in MNIST-like datasets is significantly higher than random chance. In contrast, the lack of adaptability of FFA to the different latent activity distributions hinders the network's capacity to learn, making FFA highly ineffective when not employing ReLU-like networks (see Table \ref{tab:app_rq1_acc_by_act} in Appendix \ref{app:rq1_ablation_study}). Additionally, when considering the median convergence area (last row in the table), the symmetric probability $P_s$ converges in almost half the time compared to both sigmoidal probabilities. However, this faster convergence is outweighed by the reduced maximum accuracy of models trained by Polar-FFA using this probability.

% A priori bien
Similar conclusions can be drawn from the results obtained over CIFAR-10, which are shown in Table \ref{tab:train_overall_results_cifar10}. Once again, Polar-FFA with $P_\sigma$ is the best-performing approach. Likewise, using our symmetric probability defined in Expression \eqref{eq:symmetric_prob} produces very robust models, achieving a median accuracy of $40.22$\% in comparison with the median accuracy of $12.70$\% scored by the standard FFA. When it comes to convergence speed, in this case the naive FFA algorithm scores best, followed by Polar-FFA with the symmetric probability $P_s$. Since the stopping criterion for the learning process is the same for all algorithms (early-stopping detailed in Appendix \ref{app:rq1_model_config_details}), the significantly lower median accuracy of FFA ($12.70$\%) compared to Polar-FFA ($21.46$\%) suggests that FFA undergoes premature convergence.

\begin{table}[b]
    \centering
    \caption{Maximum, median, average and minimum values of the average accuracy of the different probability functions of FFA and Polar-FFA over the CIFAR-10 dataset. We also present the amount of configurations achieving an accuracy greater than $35\%$ and the total number of configurations. Best results marked in bold.}
    \label{tab:train_overall_results_cifar10}
    
\resizebox{\columnwidth}{!}{\begin{tabular}{lccc}
\toprule[1pt]\midrule[0.3pt]
 Score & $P_{s}$ & $P_{\sigma}$     & $P_\sigma^{\textup{FFA}}$     \\
\midrule
 Maximum accuracy     &       46.39 &   \textbf{49.92} &   42.81 \\
 Median accuracy     &       \textbf{40.22} &   21.46 &   12.70  \\
 Average accuracy       &       \textbf{37.14} &   25.39 &   20.29 \\
 Minimum accuracy      &       \textbf{12.37} &    9.38 &   10.00    \\
 \midrule
\# top-35\% accuracy configurations/total &     \textbf{26/36} & 10/36 &  7/36  \\
Convergence area top-35\% configurations & 0.0651 & 0.0529 & \textbf{0.0469} \\
\midrule[0.3pt]\bottomrule[1pt]
\end{tabular}}
\end{table}

\begin{figure}[h]
    \centering
    \includegraphics[width=0.72\columnwidth]{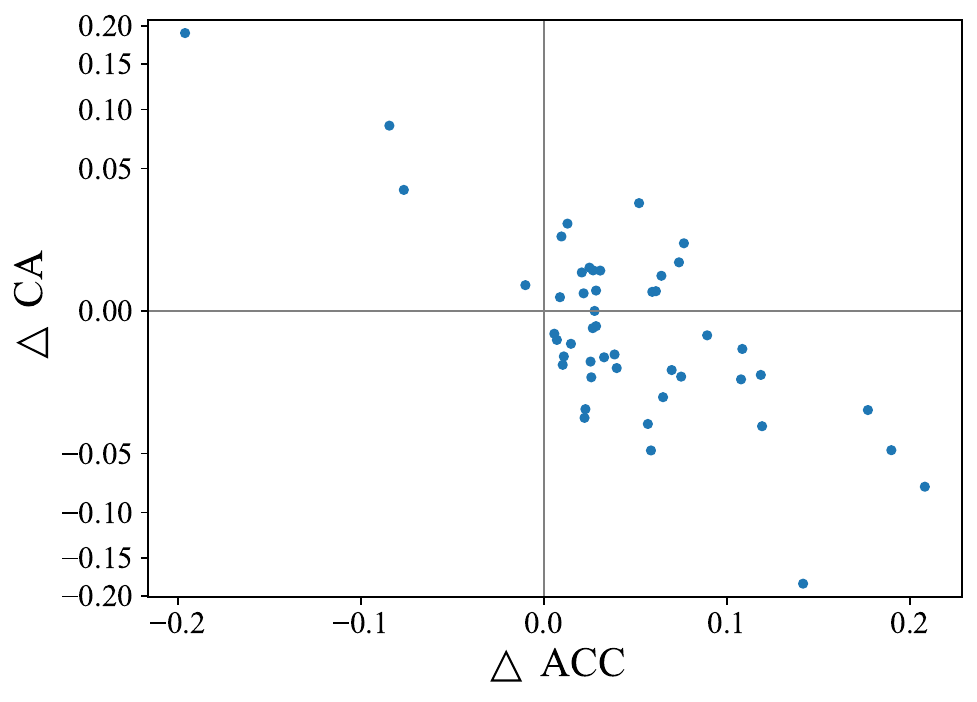}

    \caption{Distribution of the difference in convergence area $\Delta\textup{CA}$ (vertical axis) and accuracy $\Delta\textup{ACC}$ (horizontal axis) between Polar-FFA with $P_\sigma$ and FFA with $P_\sigma^{\textup{FFA}}$. The vertical axis is in square-root scale for the sake of readability. Only models with an accuracy greater than $70\%$ have been plotted, filtering those achieving fast convergence due to having suboptimal maximum accuracy.}
    \label{fig:rq1_convergence_vs_acc}
    
\end{figure}

% Convergence Area
% A priori bien
We proceed by analyzing the convergence speed of the sigmoid probability $P_\sigma$ in Polar-FFA, which closely resembles the original probability function of FFA. Figure \ref{fig:rq1_convergence_vs_acc} depicts the difference in accuracy (horizontal axis) and convergence area (vertical axis) between Polar-FFA and FFA and the configurations using $P_\sigma$ and $P_\sigma^{\textup{FFA}}$ achieving more than $70$\% accuracy in both cases. Differences in accuracy and convergence speed are predominantly inside the same quadrant, characterized by a positive difference in accuracy and a negative difference in convergence area. These results imply that incorporating neural polarization into the learning process leads to improvements in both accuracy and training speed.

%% Stability analysis
To further investigate the impact of neural configuration on the variability between training runs, we present the results of the two best performing configurations for each activation function (\texttt{ReLu}, \texttt{Tanh} and \texttt{Sigmoid}) over $10$ different seeds in Table \ref{tab:results_table_sem}. In this analysis, models trained using Polar-FFA (namely, those using $P_s$ and $P_\sigma$) exhibit a lower standard deviation across different seeds, except when using the \texttt{Sigmoid} action together with $P_\sigma$. Moreover, the results evince a clear pattern regarding the choice of the activation function in sigmoidal probability functions: models trained using \texttt{ReLu} and \texttt{Tanh} activations yield less noisy accuracy levels than models trained with \texttt{Sigmoid} activations. 

\begin{table}[h]
    \centering
    \caption{Average accuracy and standard error of the mean over $10$ runs of the best performing neural configurations with \texttt{ReLu}, \texttt{Sigmoid} and \texttt{Tanh} activations, over MNIST, Fashion-MNIST and KMNIST datasets. The highest average accuracy for each probability is highlighted in bold.}
    \label{tab:results_table_sem}
    
\resizebox{\columnwidth}{!}{\begin{tabular}{lccc}
    \toprule[1pt]\midrule[0.3pt]
 Activation, configuration & $P_{s}$ & $P_{\sigma}$     & $P_\sigma^{\textup{FFA}}$     \\  
        \midrule
\texttt{ReLu}, best conf. & $90.62 \pm 0.76$ & \textbf{92.83} $\pm$ \textbf{0.74} & \textbf{90.66} $\pm$ \textbf{0.87} \\
\texttt{ReLu}, 2$^{\textup{nd}}$ best conf. & $90.72 \pm 0.75$ & $92.81 \pm 0.73$ & $88.61 \pm 0.88$ \\
\texttt{Sigmoid}, best conf. & $91.51 \pm 0.79$ & $80.76 \pm 3.22$ & $76.01 \pm 0.79$ \\
\texttt{Sigmoid}, 2$^{\textup{nd}}$ best conf. & \textbf{91.60} $\pm$ \textbf{0.80} & $79.18 \pm 3.21$ & $71.79 \pm 0.80$ \\
\texttt{Tanh}, best conf. & $90.27 \pm 0.89$ & $91.52 \pm 0.85$ & $86.61 \pm 1.25$ \\
\texttt{Tanh}, 2$^{\textup{nd}}$ best conf. & $90.31 \pm 0.89$ & $90.93 \pm 0.93$ & $79.73 \pm 1.85$ \\
\midrule[0.3pt]\bottomrule[1pt]
\end{tabular}}
\end{table}

Out of all the results, the average accuracy remained consistent with the initial experiments, providing evidence of the robustness of FFA-like algorithms under carefully selected neural configurations. We note that experiments using the \texttt{Sigmoid} activation function give rise to a slight decrease in accuracy compared to the initially presented values in Table \ref{tab:train_overall_results} when using the sigmoidal probability function $P_\sigma$. The high variance and the accuracy decrease associated with the \texttt{Sigmoid} activation function can be attributed to a small subset of training sessions that yielded suboptimal performance, resulting in accuracy ranges between $40\%$ and $60\%$. Nevertheless, the vast majority of the training seeds produced competitive accuracy levels. In contrast, when employing the symmetric probability this effect is reversed, with the \texttt{Sigmoid} activation achieving the most accurate results. Similarly, in addition to being the most robust when working with different network configurations, this probability function also achieves the lowest deviation in accuracy between different seeds.

\subsection*{RQ2: Which insights can be obtained by analyzing the latent space induced by neural polarization?}

\begin{figure}[h]
    \centering
    \includegraphics[width=0.72\columnwidth]{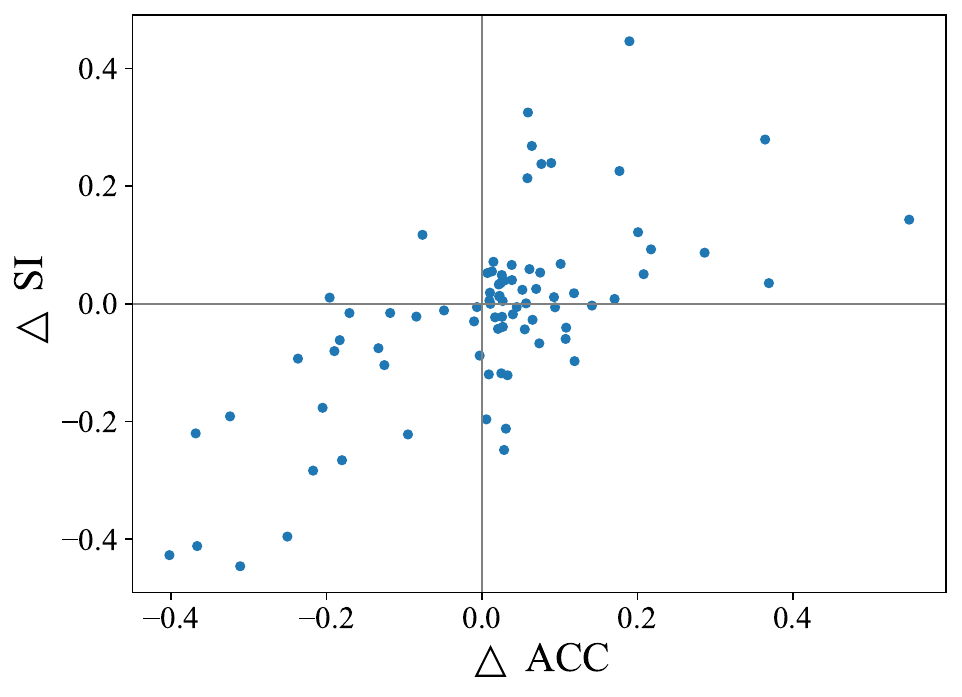}

    \caption{Distribution of the difference in the separability index $\Delta\textup{SI}$ (y-axis) and accuracy $\Delta\textup{ACC}$ (x-axis) between Polar-FFA with $P_\sigma$ and FFA with $P_\sigma^{\textup{FFA}}$. Only models with accuracy higher than $20\%$ are included, thereby filtering out near-random networks.}
    \label{fig:rq2_sep_vs_acc}
    
\end{figure}

To address this second research question, we pause at Figure \ref{fig:rq2_sep_vs_acc}, which depicts the difference in accuracy and separability between Polar-FFA and FFA for models attaining an accuracy higher than $20\%$. Conceptually, FFA is designed to maximize the separation between positive and negative samples through a contrastive learning process. Our results reveal that this goal pursued by FFA is not fulfilled in all neural configurations. For instance, a small subset of neural configurations learns to contrast between positive and negative samples by small directional perturbations driven by the embedded labels. As shown in Appendix \ref{app:rq2_non_separability}, this results in small clusters of points grouped closely, obtained from the same sample but with different embedded labels, but with the positive sample achieving a slightly greater goodness scores. The evidence of this geometrical structure points to a more diverse latent representation inherent to the algorithm, yet highly dependent on the choice of the neural configuration. Nevertheless, the major trend points towards a clear correlation between the distance between positive and negative latent spaces. 

Sparsity also reveals information about the behavior of the model. As observed by the results in Table \ref{tab:rq2_multiple_metrics}, models trained using \texttt{Tanh} as activation function result in the least sparse latent spaces, affecting the neural-level sparsity of most models except those trained using the symmetric probability function. As discussed in RQ1, these models were proven to perform most robustly, suggesting that the increased sparsity of the latent space can serve as a pivotal feature to guarantee high generalization capabilities in forward-like algorithms. In contrast, the sigmoid probability $P_\sigma$ in Polar-FFA achieves the least sparse latent spaces. This can be explained by the difference in the objectives sought by FFA and Polar-FFA. While FFA aims at reducing the overall activity in negative samples, it tends to drive down most neurons during the negative phase. On the contrary, Polar-FFA aims at maximizing the activity of the negative neural set, which involves having a high number of active neurons during the same phase. This difference in functionality creates less sparse outputs, while at the same time improves the generalization capabilities and reduces the information loss between layers. It is also important to remark that while the sparsity of Polar-FFA is lower than that of FFA, both achieve high sparsity ratios. Additionally, the highest neural usage in Table \ref{tab:rq2_multiple_metrics} signifies that a highest concentration of information is confined in a small subset of neurons for models trained using the symmetric probability $P_s$. This leads to enhanced learning dynamics arising from a higher degree of neural specialization.
\begin{table}[h]
    \centering
    \caption{Average Hoyer and neural usage of the configurations studied in RQ1. Due to the high difference between models using \texttt{Tanh} activations and those using other activation functions, we also present the average metrics of the configurations with that specific activation function excluded.}
    \label{tab:rq2_multiple_metrics}
    
    \begin{tabular}{lccc}
    \toprule[1pt]\midrule[0.3pt]
    
 Metric     & $P_{s}$ & $P_{\sigma}$     & $P_\sigma^{\textup{FFA}}$     \\
        \midrule
        Hoyer index $\textup{HI}(\bm{\ell})$ & 0.9673& 0.6430 & 0.7501 \\
        Neural usage & 0.6219& 0.2738 & 0.2074 \\
        \midrule
        Hoyer index $\textup{HI}(\bm{\ell})$ (no \texttt{Tanh}) & 0.9776& 0.7275 & 0.8766 \\
        Neural usage (no \texttt{Tanh}) & 0.8150& 0.4554 & 0.4630 \\
    \midrule[0.3pt]\bottomrule[1pt]
    \end{tabular}
\end{table}

\section{Conclusions and Future Research Lines}
\label{sec:6_concl}

This work has introduced Polar-FFA, a novel formulation of the FFA that incorporates neural polarization to enhance its learning dynamics. Our approach involves dividing each layer into positive and negative neurons, each aimed at maximizing their goodness score when presented with inputs of their respective polarity. Building upon this formulation, we propose two alternative probability functions, proven to mitigate well-known limitations of the original FFA. Through extensive experiments across a diverse set of neural configurations, including various activation and goodness functions, we provide empirical evidence of the improved generalization capabilities of Polar-FFA. Significantly, our approach consistently outperforms FFA across all datasets and nearly all neural configurations in terms of accuracy and convergence speed. Furthermore, we demonstrate its ability to learn in a broader range of neural configurations, such as models using \texttt{Sigmoid} or \texttt{Tanh} activations, where the original FFA has been proven to perform poorly. In addition, we explore the geometrical properties inherent to this extended set of configurations, showing that the higher accuracy scores produced by Polar-FFA result from its capacity to learn highly separated latent representations. Similarly, our findings highlight the positive impact that latent sparsity provides during training, leading to more robust and stable learning dynamics.

We envision two main lines to further develop the ideas explained in this work. First, we intend to advance in the study of goodness and probability functions, focusing on their emerging geometrical properties. As shown in this work, the choice of these two functions highly impacts the properties of the latent space, which could be beneficial for creating more effective networks, especially in terms of robustness against out-of-distribution data and explainability. Second, we aim to extend the heuristics from FFA to more advanced neural architectures (e.g., CNNs or Transformers), primarily by replacing the supervised negative generation method for one compatible with non-dense layers.

\begin{ack}
The authors thank the Basque Government for its funding support via the consolidated research groups MATHMODE (ref. T1256-22) and D4K (ref. IT1528-22), and the colaborative ELKARTEK project KK-2023/00012 (BEREZ-IA). E. B. Terres-Escudero is supported by a PIF research fellowship granted by the University of Deusto.
\end{ack}

\bibliography{mybibfile}

\clearpage
\newpage

\noindent \LARGE \textbf{On the Improvement of Generalization and Stability of Forward-Only Learning via Neural Polarization} 

\vspace{4mm}
\noindent \Large \textcolor{arsenic}{\textbf{Supplementary material}}
\vspace{4mm}

\appendix

\setcounter{page}{1}

\normalsize
\startcontents[sections]
\noindent\rule{\columnwidth}{1pt}

\vspace{-2mm}
\printcontents[sections]{l}{1}{\setcounter{tocdepth}{2}}

\noindent\rule{\columnwidth}{1pt}

\normalsize 
\counterwithin{figure}{section}
\counterwithin{table}{section}
\renewcommand\thefigure{\thesection\arabic{figure}}
\renewcommand\thetable{\thesection\arabic{table}}
\renewcommand\theequation{\thesection\arabic{equation}}
\setcounter{equation}{-1}

\section{Instability of Sigmoidal Probability Functions}
\label{app:prq_unbiased_sigmoid_probability}

As demonstrated by Gandhi et al. \cite{gandhi2023extending}, networks trained with FFA using bounded activation functions often exhibit suboptimal performance, sometimes even rendering models incapable of learning. This reduced learning dynamic primarily stems from the vanishing gradient behavior of the sigmoidal probability function. Given that neural networks usually initialize their weights from normal distributions, the latent vectors $\bm{\ell} \in \mathbb{R}^{n}$ resulting from the linear operation $\mathbf{W}\mathbf{x}^T$, where $\mathbf{x} \in \mathbb{R}^{m}$ is an input vector and $\mathbf{W} \in \mathbb{R}^{n \times m}$, will also follow a $0$-centered normal distribution for each coordinate. When computing the squared Euclidean norm of this vector after passing through the \texttt{Sigmoid} activation function to obtain the goodness, the resulting distribution will yield large goodness scores, with the expected value driven by the number of neurons. The exact expression is given by:
\begin{equation}
\mathbb{E}\left[\|\texttt{Sigmoid}(\bm{\ell})\|_2\right] = n \cdot \mathbb{E}[\texttt{Sigmoid}(\ell_i)] = \frac{n}{2}.
\end{equation}

Large expected goodness values only serve to degrade the learning dynamics in the traditional FFA. Given such large goodness scores, its becomes almost guaranteed that the value of the sigmoidal probability reach values close to $1$. Under this scenario, due to the behavior of the derivative of the sigmoidal function, gradient updates will become arbitrarily small, thereby producing negligible weight updates. 

To overcome this issue, a careful balance of the $\theta$ and $\alpha$ hyperparameters in the probability $P_{\sigma}$ is required. While setting $\theta$ around the distribution's expected value helps for the purpose, high variance values can still render this probability function unstable, which can be fixed by tuning $\alpha$. An initial proposal to mitigate this effect was given in \cite{gandhi2023extending}, where $\theta$ was set to the number of neurons. However, due to the difference between this value and the real expected value of the goodness, this approach would still be unable of training networks using sigmoid activations, as proven by the results therein reported. Additionally, this method does not weight the impact of the variance on the probability function, which can result in models with large variance values and the sigmoid function incapable of learning, as presented in Proposition \ref{prop:split_sigmoid_unbiased}. 

In this work, we propose the use of mean aggregation to mitigate the impact of the mean. Our hypothesis is that this strategy can reduce the correlation between the expected value of the norm and the number of neurons. However, we acknowledge that this solution still faces limitations. Due to the bounded nature of functions like \texttt{Sigmoid} or \texttt{Tanh}, the expected values of the norm can have low variance, resulting in suboptimal utilization of the probability function and inaccurate estimations of the positivity of input samples. Consequently, models employing bounded functions in FFA often achieve suboptimal performance, necessitating extensive hyperparameter tuning processes. Even when implementing such a tuning, accuracy may still be low due to distributional differences arising during model training.

% A priori bien
\section{Proof of Proposition 1}
\label{app:proposition_1_proof}

Let $z$ be the random variable obtained from the expression $G(\bm{\ell}_{\oplus})-G(\bm{\ell}_{\ominus})$, which computes the difference between the positive and negative goodness scores. Since $\bm{\ell}_{\oplus}$ and $\bm{\ell}_{\ominus}$ are independent, the same independence holds for their respective transformed values $G(\bm{\ell}_{\oplus})$ and $G(\bm{\ell}_{\ominus})$. Moreover, considering that both $G(\bm{\ell}_{\oplus})$ and $G(\bm{\ell}_{\ominus})$ originate from transformations of weights drawn from the same distribution, they share identical distributions, implying equal mean values. This equality implies that the mean value of $z$ will be given by:
\begin{equation}
    \mathbb{E}[z] = \mathbb{E}\left[G(f(\mathbf{W}_{\oplus}\mathbf{x}^T)) - G(f(\mathbf{W}_{\ominus}\mathbf{x}^T))\right] = 0
\end{equation}

Additionally, it is clear that the distribution of the variable $z$ is symmetric, as any point $z \in \mathbb{R}$ satisfies that $g(z) = g(-z)$, where $g$ represents the probability density function of $z$. Given this property and the fact that the function $\sigma$ satisfies that $\sigma(-x) = 1-\sigma(x)$, we can easily verify that $\mathbb{E}[P_\sigma(z)] =  0.5$.

Given that the derivative of a sigmoidal function is $\sigma'(x) = \sigma(x)(1-\sigma(x))$, we can express the expected value of the derivative as:
\begin{equation}
\label{eq:base_sigmoid_expression}
    \mathbb{E}\left[\frac{\partial P_\sigma(z)}{\partial G(\bm{\ell}_{\oplus})}  \right] = \mathbb{E}\left[P_\sigma(z)(1- P_\sigma(z))\right].
\end{equation}

Using the previously stated fact that $\mathbb{E}[z] = 0.5$, we can manipulate the previous expression to obtain the following simplification:
\begin{equation*}
    \frac{1}{4}\mbox{ -- }\mathbb{E} \left[ \frac{1}{4}\mbox{ -- }P_\sigma(z) + (P_\sigma(z))^2\right] = 
    \frac{1}{4}\mbox{ -- }\mathbb{E} \left[ \left(\mathbb{E}[P_\sigma(z)] \mbox{ -- } P_\sigma(z)\right)^2\right]
\end{equation*}

In the last equation, the expected value on the right side is by definition the variance of the sigmoidal activity. Therefore, we can reformulate the expression into the desired statement, thereby completing the proof:
\begin{equation}
    \mathbb{E}\left[\frac{\partial P_\sigma(z)}{\partial G(\bm{\ell}_{\oplus})}  \right] = \frac{1}{4}-\textup{Var} \left[ P_\sigma(z)\right] \geq 0.
\end{equation}

The second statement of the proposition aims to provide a lower bound for the previously mentioned expression. This lower bound is computed by employing a function that covers the original sigmoid. While there exist other covering functions that more closely approximate the original sigmoid, offering stricter lower bounds, for the purpose of this proof we limit ourselves to acknowledging the existence of such bounds. The chosen function is $x^2+0.25$, which can be shown to exceed $\sigma^2(x)$ over the range of real numbers. Using properties of the expected value, we prove that:
\begin{equation}
    \label{eq:inequality_sigma}
    \mathbb{E}[P_{\sigma}^2(z)] \leq 0.25 + \mathbb{E}[z^2] = 0.25 + \textup{Var}[z^2].
\end{equation}

By expanding Equation \eqref{eq:base_sigmoid_expression}, we obtain:
\begin{equation}
    \mathbb{E}\left[\frac{\partial P_\sigma(z)}{\partial G(\bm{\ell}_{\oplus})}  \right] = \underbrace{\mathbb{E}\left[P_\sigma(z)\right]}_{=0.5}- \mathbb{E}\left[P_\sigma(z))\right],
\end{equation}
from which, by substituting the value of $\mathbb{E}\left[P_\sigma(z))\right]$ using the inequality in Equation \eqref{eq:inequality_sigma}, we obtain the desired formula:
\begin{equation}
    \mathbb{E}\left[\frac{\partial P_\sigma(z) }{\partial G(\bm{\ell}_{\oplus})} \right] \geq 0.25 - \textup{Var}\left[z^2\right].
\end{equation}

% A priori bien
\setcounter{equation}{0}
\section{Proof of Proposition 2}
\label{app:proposition_2_proof}

To prove the first statement of this proposition, we will verify that scaling the value of the goodness scores is equivalent to transforming the value of $\epsilon$. If the given transformation of $\epsilon$ is smaller than the values of the goodness, we can verify that the function is approximately equivalent to completely removing the $\epsilon$ value. To do so, let $\gamma$ be a non-negative scaling factor. Then we have that:
\begin{equation}
    \frac{\gamma G(\bm{\ell}_{\oplus}) +\epsilon}{\gamma G(\bm{\ell}_{\oplus})+\gamma G(\bm{\ell}_{\ominus})+2\epsilon} = \frac{ G(\bm{\ell}_{\oplus}) +\epsilon\gamma^{-1}}{G(\bm{\ell}_{\oplus})+G(\bm{\ell}_{\ominus})+2\epsilon\gamma^{-1}}.
\end{equation}

Given the constraint $\gamma G(\bm{\ell}_{\oplus}) \gg \epsilon$, we have that $ G(\bm{\ell}_{\oplus}) \gg \epsilon\gamma^{-1}$, and, since the value of $\epsilon\gamma^{-1}$ is numerically negligible when compared to goodness scores, it follows that:
\begin{equation}
    \frac{ G(\bm{\ell}_{\oplus}) +\epsilon\gamma^{-1}}{G(\bm{\ell}_{\oplus})+G(\bm{\ell}_{\ominus})+2\epsilon\gamma^{-1}} \approx \frac{ G(\bm{\ell}_{\oplus})}{G(\bm{\ell}_{\oplus}) +G(\bm{\ell}_{\ominus})}
\end{equation}

The second statement of the proposition is deduced from a direct computation of the expression's derivative. Since the value of $\epsilon$ does not affect the expression significantly when $G(\bm{\ell})$ is much larger than $\epsilon$, we will omit it for the remainder of the proof. Therefore, the derivative is expressed as:
\begin{equation}
    \frac{\partial P_s(G(\bm{\ell}_{\oplus}), G(\bm{\ell}_{\ominus}))}{\partial G(\bm{\ell}_{\oplus})}  = \frac{G(\bm{\ell}_{\ominus})}{G(\bm{\ell}_{\oplus})}\frac{1}{G(\bm{\ell}_{\oplus}) + G(\bm{\ell}_{\ominus})},
\end{equation}
which converges to zero under two conditions: i) when the sum of the goodness values approaches infinity, or ii) when the ratio between negative and positive goodness values tends to zero. Since we assumed a upper and lower bounded sum of goodness values, the derivative's value is predominantly influenced by the ratio of goodness values. Consequently, this implies that:
\begin{equation}
    \frac{\partial P_s(G(\bm{\ell}_{\oplus}), G(\bm{\ell}_{\ominus}))}{\partial G(\bm{\ell}_{\oplus})}  = \mathcal{O}\left(\frac{G(\bm{\ell}_{\ominus})}{G(\bm{\ell}_{\oplus})}\right),
\end{equation}
with $\mathcal{O}(\cdot)$ denoting asymptotic order of complexity.

\section{Experimental Setup: Additional Information}
\label{app:rq1_model_config_details}

This appendix provides several additional details of the experimental setup, such as the choice of hyperparameter values and the total set of neural configuration chosen for the experiments in RQ1.

% A priori bien
\paragraph{Hyperparameter values} The hyperparameter values for the sigmoidal function in FFA were retrieved from those used in the original work of Hinton \cite{hinton2022forward}: $\theta = 2$ and $\alpha=1$. However, considering the arguments presented in Appendix \ref{app:prq_unbiased_sigmoid_probability}, we opt to use different hyperparameter values for models trained using the \texttt{Sigmoid} activation, as they provide better-than-random accuracy in a small set of experiments. For neural configurations employing the $L_2$ norm or its square variation, we use $\theta = 0.2$ and $\alpha = 5$. Configurations using the $L_1$ norm consider $\theta = 0.4$ and $\alpha = 2.5$. The parameters for the Polar-FFA's version of the sigmoid probability ($P_\sigma$) remained consistent with the original $\theta=2$ and $\alpha=1$ values for all experiments. The $\epsilon$ value of the symmetric probability $P_s$ was set to $10^{-6}$ for all experiments to ensure numerical stability. No additional hyperparameter tuning was considered. As mentioned in Section \ref{sec:3_characterization}, the division of positive and negative neurons followed a 1-to-1 relationship, meaning that each layer had the same number of positive and negative neurons.

\paragraph{Dataset Configuration} To maintain consistency across experiments, all datasets are normalized by standardizing their values. Similarly, we employ the original train/validation/test partitions of all datasets across all experiments. No data augmentation techniques were used.

% A priori bien
\paragraph{Early Stopping on CIFAR-10} To mitigate computational overhead, an early stopping strategy was applied during the experiments related to the CIFAR-10 dataset. Models that did not show improvement in accuracy for more than 10 epochs were stopped before reaching their $T=100$ epoch limit. This strategy has been used due to the large amount of models showing suboptimum or even close-to-random performance.

% A priori bien
\paragraph{Neural Configurations} To provide experimental results over a comprehensive set of neural configurations, we trained models using combinations of the activation, goodness, and probability functions proposed in this appendix. This approach resulted in $108$ different configurations being employed for each dataset. The set of probability functions comprises the original Sigmoid probability of FFA, the Sigmoid Probability of Polar-FFA, and the Symmetric Probability of Polar-FFA. The choice of activation functions was made to cover activations with different behaviors. For this instance, we selected: the \texttt{ReLU} function, due to its unbounded and non-negative behavior; the \texttt{Sigmoid} function, due to its bounded and non-negative behavior; and the hyperbolic tangent function \texttt{Tanh}, as its range is not limited to positive values. This selection is presented in Table \ref{tab:app_rq1_training_acti_prob_config}.
\begin{table}[h!]
    \centering
    \caption{Activation and probability functions chosen to build the set of neural configurations for the training experiments aimed to answer RQ1.}\label{tab:app_rq1_training_acti_prob_config}
    
    \begin{tabular}{cc}
        \toprule[1pt]\midrule[0.3pt]
         Activation function & Probability function  \\
          \midrule
         \texttt{ReLU}, \texttt{Sigmoid}, \texttt{Tanh} & $P_{\sigma}^{\textup{FFA}}$,  $P_{\sigma}$, $P_{s}$  \\
         \midrule[0.3pt]\bottomrule[1pt]
    \end{tabular}
\end{table}

% A priori bien
To explore a diverse set of goodness functions, we divide its functionality into three distinct components: i) the employed norm, which measures how the latent vector is evaluated; ii) the aggregation method of the norm, which can vary between the sum or the mean of the elements; and iii) the lateral inhibition mechanism, which limits the number of active neurons in the latent vector, thereby restricting the number of modified synapses during learning. While Hinton advocated for the use of the squared Euclidean norm in the original approach, mainly due to the simplicity of the gradient, we investigate the use of two additional norm functions: the $\|\cdot\|_2$ norm and the $\|\cdot\|_1$ norm. Additionally, we analyze the different behavior of the models when employing a mean-based and a sum-based aggregation method for the norm. This choice is mainly motivated by the arguments presented in Appendix \ref{app:prq_unbiased_sigmoid_probability} regarding possible solutions to the vanishing gradient that arises when employing a \texttt{Sigmoid} activation. In this case, replacing the sum-based with a mean-based norm implies reducing the variance of the sigmoidal distribution, which mitigates the vanishing effect in some cases.

% A priori bien
Finally, we investigate the impact of incorporating a lateral inhibition mechanism into the latent vector. These methods were introduced to increase sparsity in the latent vectors, thereby regulating activity and limiting the number of weights updated at each forward pass. This was achieved by implementing a \emph{Winner-Takes-All} (WTA) dynamic using a top-k selection function. Specifically, this function resets to zero all elements in a vector that are not among the $k$ most active elements. For all experiments, the value for $k$ was set to $15$ to ensure that only a small subset of neurons was active. It is important to remark that the lateral inhibition scheme is only directly applied to the goodness function. However, in order to limit the information loss between layers, the non-inhibited latent vector is passed to subsequent layers. A summarized version of the different components of the goodness function is given in Table \ref{tab:app_rq1_training_goodness_config}. 
\begin{table}[h]
    \centering
    \caption{List of the goodness components that are considered in the training experiments related to RQ1. Each goodness function was composed by employing a norm, an aggregation method and a lateral inhibition mechanism. Each column lists all the methods for each component.}
    \label{tab:app_rq1_training_goodness_config}
    
    \resizebox{\columnwidth}{!}{\begin{tabular}{ccccc}
        \toprule[1pt]\midrule[0.3pt]
        \multicolumn{3}{c}{Goodness configuration} \\ \midrule
          Norm function & Aggregation method  & Lateral inhibition\\
          \midrule
          $\|\bm{\ell}\|_2^2$, $\|\bm{\ell}\|_2$, $\|\bm{\ell}\|_1$ & Mean, sum & No inhibition, WTA inhibition \\
         \midrule[0.3pt]\bottomrule[1pt]
    \end{tabular}}
\end{table}

\section{Effect on the Ratio of Positive and Negative Neurons}
\label{app:ratio_neurons}
%% A priori bien
This appendix provides further analysis of the performance obtained when employing different relations of positive to negative neurons. To provide a systematic analysis of distinct scenarios, we examined both balanced and highly disproportionate polarity distributions. Specifically, we analyzed networks with the following percentages of positive neurons at each layer: 5\%, 10\%, 25\%, 50\%, 75\%, 90\%, and 95\%. We trained multiple networks using the same neural configurations described in Appendix \ref{app:rq1_model_config_details}, except for those using WTA mechanics. These experiments were constrained to grayscale datasets: MNIST, Fashion-MNIST, and K-MNIST. All networks were trained for 10 epochs using the same set of hyperparameters as in the RQ1 experiments. Since the only metric of interest in this appendix is the variation in accuracy across the different positive-to-negative splits, we normalized the results of each configuration relative to their mean accuracy.

\begin{figure}[h]
    \centering
    \includegraphics[width=\linewidth]{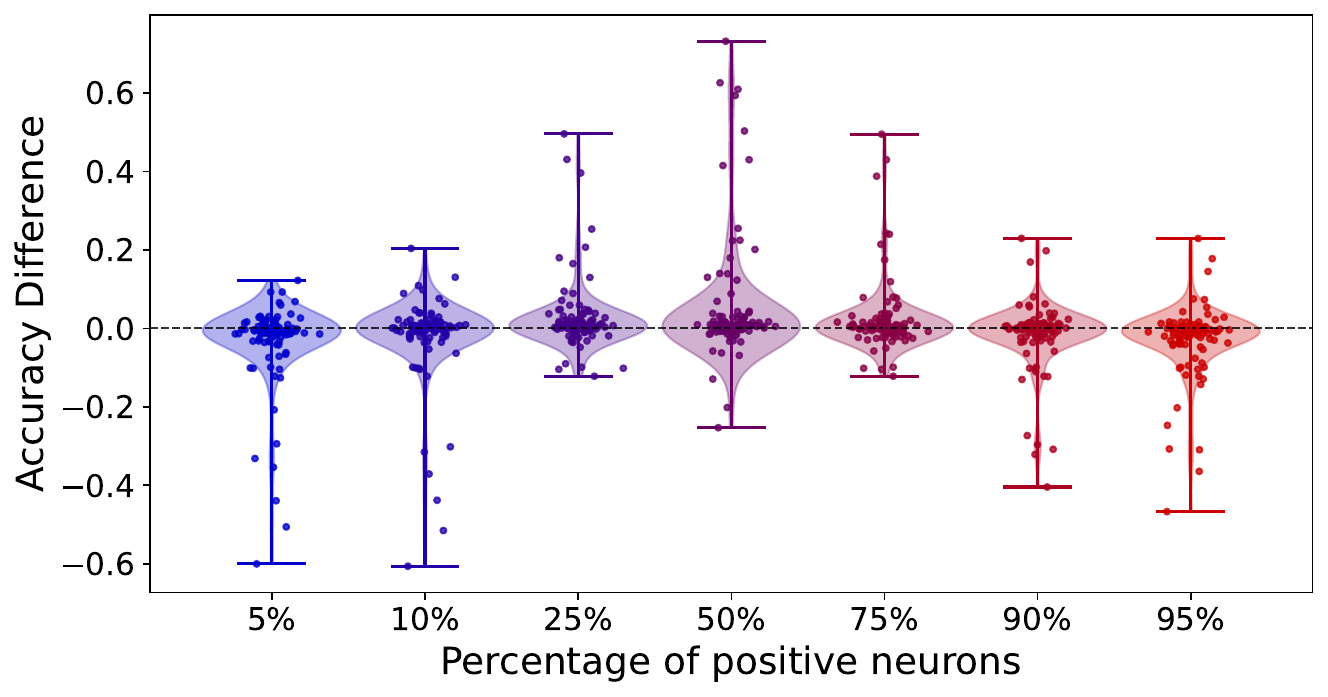}
    \caption{Distribution in the difference in accuracy between the distinct percentages of positive to negative neurons to the mean accuracy of the different percentages.}
    \label{fig:acc_diff}
    
\end{figure}

The results of these experiments are depicted in Figure \ref{fig:acc_diff}. Overall, they demonstrate that most neural configurations achieve comparable accuracy, with only a small subset of outliers exhibiting a statistically significant difference in accuracy. This finding suggests that Polar-FFA networks possess self-regulatory dynamics, where the activation strength of the two polarity sets adjusts to surpass the other set when presented with inputs of their respective polarity. However, when examining the general behavior of the outliers, a clear tendency emerges: they perform better with a balanced distribution of positive and negative neurons, while showing reduced accuracies at the extremes of polarity distribution.

From an experimental standpoint, given the observed results, it appears clear that the best polarity distribution within the layers is achieved with a near-balanced configuration. Consequently, this configuration was used in the experiments conducted for both RQ1 and RQ2.

\begin{figure}[h]
    \centering
    \includegraphics[width=\linewidth]{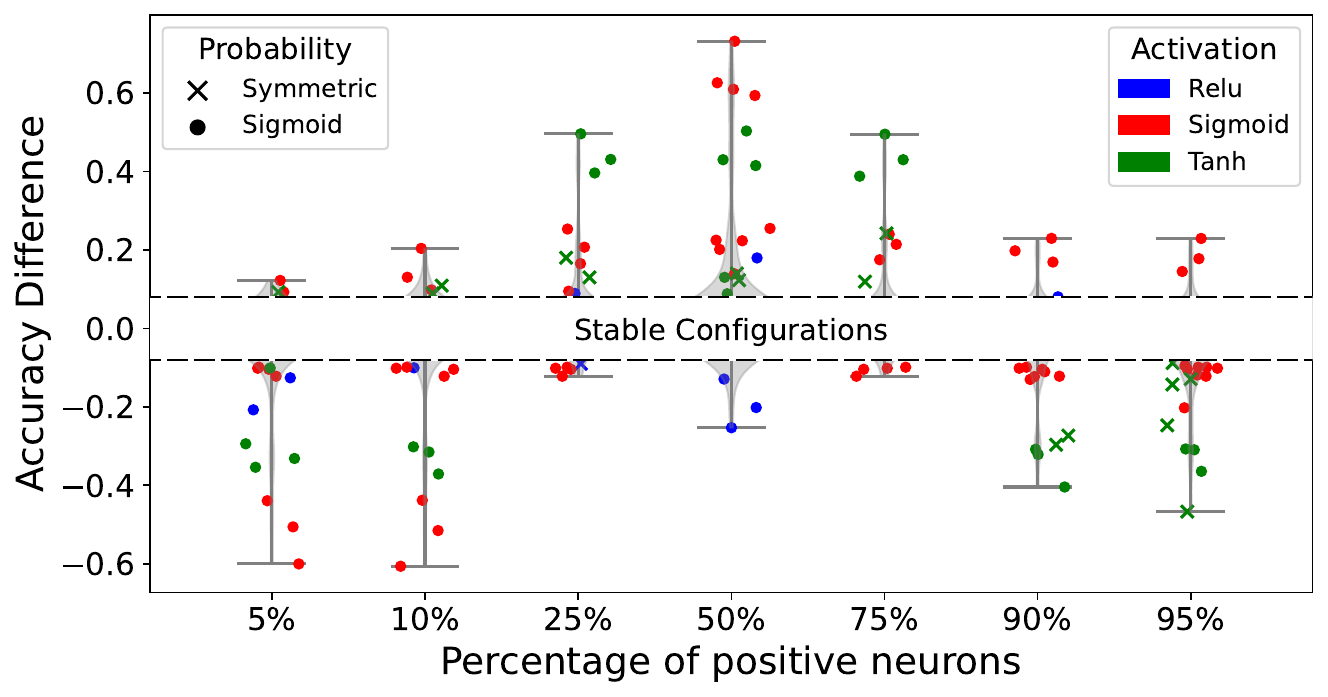}
    \caption{Plot showing the accuracy difference of the outlying elements, categorized based on the probability function and the activation function used. Points with less than 0.08 absolute accuracy difference, denoted as \textit{stable configuration band}, have been removed to improve clarity.}
    \label{fig:acc_diff2}
    \vspace{3mm}
\end{figure}

However, to gain further insights into this set of outliers, we present evidence on the circumstances under which these outliers emerge. Figure \ref{fig:acc_diff2} provides a disaggregated view of the results, showing the activation and probability functions used during training. For clarity, we exclude the set of \textit{stable configurations}, as they do not provide relevant information regarding the outliers. This band of configurations ranges within an accuracy difference of $[-0.08, 0.08]$, encompassing 85\% of the most stable experimental results.

\paragraph{Effects of the Probability Function.} Upon early inspection, the Symmetric probability function appears to produce more stable results, with only a small set of outliers exhibiting a characteristic behavior: all result from using the \texttt{Tanh} activation function with a high percentage of positive neurons. In this scenario, the model seems incapable of balancing the high positivity of the large neural dataset with the reduced negative neural set. We believe this asymmetry in the results arises from the wider negative data distribution, which arises from the combination of input samples with all non-corresponding labels, compared to the sharper distribution of positive samples. In contrast, most outliers appear when employing the Sigmoidal probability function, especially when using bounded activation functions. Unlike the Symmetric probability, these outliers show a sharper decrease in accuracy when the positive neural set is smaller than the negative set. This effect can be attributed to the inability of the positive neural set to achieve a high enough goodness score to push the probability function towards non-vanishing output ranges.

\paragraph{Effects of the Activation Function.} Out of the three functions, \texttt{ReLU} demonstrates the most competitive performance, with its few outliers highly concentrated in balanced polarity distributions, showing only a small accuracy difference of \(\pm 0.20\). \texttt{Tanh} exhibits a clear preference for near-equilibrium ratios of positive to negative neurons, often resulting in a negative accuracy difference when the positive-to-negative neural distributions are imbalanced. Similarly, \texttt{Sigmoid} performs better with even polarity distributions, slightly improving its results in configurations where the positive neurons outnumber the negative ones. In cases with a very low percentage of positive neurons, \texttt{Sigmoid} yields reduced accuracy, which can be attributed to the positive neurons' inability to achieve high goodness scores necessary to push the probability function into stable ranges. Among the two bounded functions, \texttt{Tanh} shows a less drastic drop in accuracy compared to \texttt{Sigmoid}, with most outliers remaining within the \(\pm 0.40\) range, while \texttt{Sigmoid} reaches the \(\pm 0.60\) range.

\section{Additional Results for RQ1: goodness, probability functions and activations}
\label{app:rq1_ablation_study}

In this appendix we present additional results for answering RQ1, broken down based on different configuration parameters. Table \ref{tab:app_rq1_acc_by_goodness} shows the average accuracy scores of the different goodness functions over MNIST-like datasets, whereas Table \ref{tab:app_rq1_acc_by_act} presents the average accuracy focusing solely on the activation function. Finally, the complete set of results for each neural combination and dataset is presented in Table \ref{tab:total_accuracies} (shown in the next page).
\begin{table}[h]
    \centering
    \caption{Mean accuracy of each goodness configurations averaged over \texttt{ReLU}, \texttt{Sigmoid} and \texttt{Tanh} activated models and all MNIST-like datasets. The highest average accuracy is highlighted in bold.}
    \label{tab:app_rq1_acc_by_goodness}
    
    \begin{tabular}{cccccc}
    \toprule[1pt]\midrule[0.3pt]
    Norm   & Aggregation & Inhibition & $P_{s}$ & $P_{\sigma}$ & $P_\sigma^{\textup{FFA}}$     \\
    \midrule
    \multirow{4}{*}{$\|\bm{\ell}\|_2^2$}
            & \multirow{2}{*}{Sum}  & --     & 83.80     &  81.28   &  59.08    \\
            &                       & WTA  & 90.75    &  91.67   &  60.89    \\
            \cmidrule{2-6}
            & \multirow{2}{*}{Mean} & --     & 84.05    &  68.37   &  66.26     \\
            &                       & WTA  & \textbf{90.76}    &  72.56   &  41.97      \\
            \midrule
        \multirow{4}{*}{$\|\bm{\ell}\|_2$}
            & \multirow{2}{*}{Sum}  & --     & 83.41     &  87.14   &  55.97    \\
            &                       & WTA  & 89.53     &  \textbf{91.96}   &  60.49    \\
            \cmidrule{2-6}
            & \multirow{2}{*}{Mean} & --     & 82.87     &  61.88   &  63.99    \\
            &                       & WTA  & 89.29     &  61.30   &  34.86    \\
            \midrule
        \multirow{4}{*}{$\|\bm{\ell}\|_1$}
            & \multirow{2}{*}{Sum}  & --     & 81.34     &  90.33   &  50.01  \\
            &                       & WTA  & 88.14     &  90.16   &  \textbf{83.45}  \\
            \cmidrule{2-6}
            & \multirow{2}{*}{Mean} & --     & 81.21     &  67.63   &  65.15     \\
            &                       & WTA  & 87.91     &  67.89   &  55.99    \\
         \midrule[0.3pt]\bottomrule[1pt]
    \end{tabular}
\end{table}

One straightforward observation from the results of the different neural configurations is a notable increase in accuracy when employing WTA inhibition dynamics on models using the symmetric probability $P_s$. However, this effect seems to be more dependent on the aggregation strategy when using the other probability functions. Models tend to achieve higher accuracy levels when employing a sum-based aggregation, while this score drops when a mean-based aggregation is used. We hypothesize that this effect relates to the activity bounds resulting from each neural configuration. When reducing the activity with WTA dynamics on mean-based goodness functions, it yields low goodness scores, restricting the behavior of the probability and thereby leading to degraded accuracy. Conversely, the effect is reversed in the sum-based score, where WTA reduces the variance of high-valued goodness scores. Surprisingly, the average accuracy over the three activation functions in FFA appears to achieve maximal accuracy when employing the $L_1$ norm, in contrast with the original approach which advocated for the usage of the squared Euclidean norm. Nevertheless, all models achieving the highest accuracy, which were used to produce the results in Table \ref{tab:results_table_sem}, employed the $\|\cdot\|_2$ norm.
\begin{table}[h]
    \centering
    \caption{Mean accuracy of each activation function on each probability function averaged over the MNIST, KMNIST, Fashion-MNIST and CIFAR-10 datasets.}
    \label{tab:app_rq1_acc_by_act}
    
    \begin{tabular}{cccc}
        	\toprule[1pt]\midrule[0.3pt]
       Activation     & $P_{s}$ & $P_{\sigma}$     & $P_\sigma^{\textup{FFA}}$     \\
        \midrule
            \texttt{ReLU} &   87.76     &  \textbf{87.04}   &  \textbf{84.04}    \\
            \texttt{Sigmoid} &   \textbf{90.78}     &  69.41   &  40.45    \\
            \texttt{Tanh} &   79.72     &  76.59   &  50.03    \\
         \midrule[0.3pt]\bottomrule[1pt]
    \end{tabular}
\end{table}

When it comes to the results discussed in Section \ref{sec:subsec_rq1}, we observe that \texttt{ReLU} achieves the highest generalization score in models trained using  $P_{\sigma}$ (Polar-FFA) or $P_{\sigma}^{\textup{FFA}}$ (FFA). Both the \texttt{Sigmoid} and the \texttt{Tanh} activation functions are found to achieve lower accuracy scores. In contrast, when examining the results of the $P_s$ probability, the opposite appears to hold, with \texttt{Sigmoid} activation emerging as the best-performing activation function. Additionally, as hinted by previous results, the difference between the average accuracy of the different activation functions is minimized when using the symmetric probability $P_s$, while $P_{\sigma}^{\textup{FFA}}$ produces the highest variance in the reported performance figures.
\begin{table*}[b]
    \centering
    \caption{Highest test accuracy obtained for each experiments employing the different neural combinations.}
    \label{tab:total_accuracies}
    
    \resizebox{1.75\columnwidth}{!}{%
    \begin{tabular}{ccccccccccccccc}

        	\toprule[1pt]\midrule[0.3pt]
         \multicolumn{15}{c}{MNIST Dataset} \\
        \midrule[0.3pt]
       
         Norm & Aggregation & Inhibition     & & \multicolumn{3}{c}{$P_{s}$} & & \multicolumn{3}{c}{$P_{\sigma}$}   &  & \multicolumn{3}{c}{$P_\sigma^{\textup{FFA}}$}     \\
          \midrule[0.3pt]
       &  &  & & \texttt{Sigmoid} & \texttt{ReLU} & \texttt{Tanh}  & & \texttt{Sigmoid} & \texttt{ReLU} & \texttt{Tanh}   & & \texttt{Sigmoid} & \texttt{ReLU} & \texttt{Tanh}    \\ 
    \midrule[0.3pt]
        
        \multirow{4}{*}{$\|\bm{\ell}\|_2^2$}
            & \multirow{2}{*}{Sum}  & -     && 97.30 & 96.31 & 79.30    &&  43.70 & 87.62 & 97.61   & & 9.80 & 96.06 & 91.57    \\
            &                       & WTA  && 97.27 & 95.79 & 96.94    &&  98.32 & 97.37 & 97.98   & & 9.80 & 95.91 & 95.86    \\
            & \multirow{2}{*}{Mean} & -     & &97.41 & 96.52 & 80.37    &&  52.84 & 96.87 & 75.25   & & 89.64 & 95.99 & 14.02    \\
            &                       & WTA  & &97.58 & 96.15 & 97.01    &&  64.15 & 97.07 & 86.27   & & 11.56 & 94.10 & 7.60    \\
            \midrule
        \multirow{4}{*}{$\|\bm{\ell}\|_2$}
            & \multirow{2}{*}{Sum}  & -     & &96.32 & 91.20 & 80.49     & & 96.53 & 98.26 & 96.19   &&  9.80 & 95.68 & 75.40   \\
            &                       & WTA  && 96.44 & 91.99 & 96.60     &&  95.94 & 98.15 & 96.76   &&  79.21 & 97.33 & 93.91   \\
            & \multirow{2}{*}{Mean} & -     & &96.34 & 91.21 & 80.89     & & 52.64 & 93.20 & 76.29   &&  88.08 & 93.77 & 9.94   \\
            &                       & WTA  & &96.53 & 91.61 & 96.64     & & 65.39 & 93.17 & 75.02   &&  90.56 & 91.39 & 4.07   \\
            \midrule
        \multirow{4}{*}{$\|\bm{\ell}\|_1$}
            & \multirow{2}{*}{Sum}  & -     & &97.23 & 95.23 & 82.36     &&  97.85 & 73.82 & 97.21   &&  9.80 & 93.42 & 88.24   \\
            &                       & WTA  && 97.13 & 94.51 & 96.58     &&  98.10 & 98.25 & 97.59   &&  9.80 & 95.00 & 93.72   \\
            & \multirow{2}{*}{Mean} & -     & &97.15 & 94.48 & 82.04     &&  43.94 & 91.42 & 73.69   &&  85.18 & 90.88 & 12.76   \\
            &                       & WTA  & &96.77 & 93.43 & 96.59     &&  69.84 & 85.05 & 60.88   &&  9.75 & 75.35 & 7.61   \\
            \midrule[0.3pt]\midrule[0.3pt]
        
\multicolumn{15}{c}{K-MNIST Dataset} \\
         \midrule[0.3pt]
           Norm & Aggregation & Inhibition     & & \multicolumn{3}{c}{$P_{s}$} & & \multicolumn{3}{c}{$P_{\sigma}$}   &  & \multicolumn{3}{c}{$P_\sigma^{\textup{FFA}}$}     \\
           \midrule[0.3pt]
           &  &  & & \texttt{Sigmoid} & \texttt{ReLU} & \texttt{Tanh}  & & \texttt{Sigmoid} & \texttt{ReLU} & \texttt{Tanh}   & & \texttt{Sigmoid} & \texttt{ReLU} & \texttt{Tanh}    \\
        \midrule[0.3pt]
        \multirow{4}{*}{$\|\bm{\ell}\|_2^2$}
            & \multirow{2}{*}{Sum}  & -     && 87.98 & 87.89 & 52.20    &&  89.59 & 87.61 & 87.64   & & 10.00 & 81.84 & 67.27    \\
            &                       & WTA  && 89.69 & 87.86 & 88.01    &&  90.58 & 87.99 & 89.80   & & 10.00 & 77.15 & 79.10    \\
            & \multirow{2}{*}{Mean} & -     & &87.80 & 88.54 & 52.99    &&  39.02 & 90.35 & 42.18   & & 63.52 & 83.72 & 39.76    \\
            &                       & WTA  & &89.20 & 88.10 & 87.81    &&  37.29 & 86.97 & 67.54   & & 8.60 & 79.29 & 10.39    \\
            \midrule
        \multirow{4}{*}{$\|\bm{\ell}\|_2$}
            & \multirow{2}{*}{Sum}  & -     & &86.77 & 79.13 & 51.53     & & 84.51 & 91.50 & 82.89   &&  10.00 & 84.59 & 46.12   \\
            &                       & WTA  && 88.83 & 76.94 & 86.04     &&  82.54 & 91.57 & 85.25   &&  59.80 & 87.60 & 77.88   \\
            & \multirow{2}{*}{Mean} & -     & &85.85 & 77.29 & 52.05     & & 58.28 & 80.86 & 37.58   &&  59.59 & 74.21 & 42.41   \\
            &                       & WTA  & &87.58 & 76.59 & 85.86     & & 35.01 & 80.59 & 53.95   &&  67.50 & 72.71 & 9.21   \\
            \midrule
        \multirow{4}{*}{$\|\bm{\ell}\|_1$}
            & \multirow{2}{*}{Sum}  & -     & &88.05 & 86.63 & 51.56     &&  88.35 & 89.67 & 87.58   &&  10.00 & 71.90 & 58.80   \\
            &                       & WTA  && 89.09 & 82.41 & 86.15     &&  90.41 & 91.07 & 88.86   &&  10.00 & 79.14 & 77.02   \\
            & \multirow{2}{*}{Mean} & -     & &86.76 & 84.95 & 52.57     &&  23.78 & 76.45 & 40.04   &&  61.43 & 66.66 & 40.66   \\
            &                       & WTA  & &87.73 & 84.25 & 86.26     &&  29.60 & 69.41 & 48.30   &&  10.30 & 59.58 & 9.30   \\
            \midrule[0.3pt]\midrule[0.3pt]
        
\multicolumn{15}{c}{Fashion-MNIST Dataset} \\
         \midrule[0.3pt]
                Norm & Aggregation & Inhibition     & & \multicolumn{3}{c}{$P_{s}$} & & \multicolumn{3}{c}{$P_{\sigma}$}   &  & \multicolumn{3}{c}{$P_\sigma^{\textup{FFA}}$}     \\
        \midrule[0.3pt] 
        &  &  & & \texttt{Sigmoid} & \texttt{ReLU} & \texttt{Tanh}  & & \texttt{Sigmoid} & \texttt{ReLU} & \texttt{Tanh}   & & \texttt{Sigmoid} & \texttt{ReLU} & \texttt{Tanh}    \\
        \midrule[0.3pt]
        \multirow{4}{*}{$\|\bm{\ell}\|_2^2$}
            & \multirow{2}{*}{Sum}  & -     && 87.83 & 87.07 & 78.29    &&  87.95 & 62.77 & 87.04   & & 10.00 & 84.50 & 80.69    \\
            &                       & WTA  && 87.85 & 87.30 & 86.07    &&  88.74 & 87.02 & 87.22   & & 10.00 & 85.62 & 84.59    \\
            & \multirow{2}{*}{Mean} & -     & &87.86 & 86.95 & 78.00    &&  61.42 & 87.76 & 69.67   & & 76.28 & 85.54 & 47.88    \\
            &                       & WTA  & &87.84 & 86.85 & 86.28    &&  61.58 & 87.05 & 65.12   & & 71.21 & 83.86 & 11.08    \\
            \midrule
        \multirow{4}{*}{$\|\bm{\ell}\|_2$}
            & \multirow{2}{*}{Sum}  & -     & &87.00 & 84.91 & 74.68     & & 86.97 & 88.92 & 87.16   &&  10.00 & 86.27 & 32.22   \\
            &                       & WTA  && 87.47 & 83.50 & 85.42     &&  86.20 & 88.45 & 86.59   &&  84.04 & 87.36 & 83.90   \\
            & \multirow{2}{*}{Mean} & -     & &86.70 & 84.75 & 75.81     & & 58.99 & 85.09 & 65.70   &&  72.57 & 83.92 & 61.86   \\
            &                       & WTA  & &86.95 & 83.56 & 85.89     & & 59.43 & 84.83 & 63.59   &&  77.41 & 81.40 & 9.65   \\
            \midrule
        \multirow{4}{*}{$\|\bm{\ell}\|_1$}
            & \multirow{2}{*}{Sum}  & -     & &87.96 & 87.29 & 74.37     &&  88.10 & 75.47 & 86.21   &&  10.00 & 83.11 & 78.46   \\
            &                       & WTA  && 87.72 & 86.23 & 85.99     &&  88.55 & 88.51 & 86.33   &&  10.00 & 86.05 & 83.66   \\
            & \multirow{2}{*}{Mean} & -     & &87.07 & 86.58 & 74.23     &&  53.56 & 83.73 & 70.27   &&  72.20 & 81.41 & 64.76   \\
            &                       & WTA  & &87.07 & 85.48 & 86.03     &&  49.08 & 79.68 & 59.86   &&  68.94 & 63.04 & 9.84   \\
            \midrule[0.3pt]\midrule[0.3pt]
            \multicolumn{15}{c}{CIFAR-10 Dataset} \\
         \midrule[0.3pt]
                Norm & Aggregation & Inhibition     & & \multicolumn{3}{c}{$P_{s}$} & & \multicolumn{3}{c}{$P_{\sigma}$}   &  & \multicolumn{3}{c}{$P_\sigma^{\textup{FFA}}$}     \\
        \midrule[0.3pt] 
        &  &  & & \texttt{Sigmoid} & \texttt{ReLU} & \texttt{Tanh}  & & \texttt{Sigmoid} & \texttt{ReLU} & \texttt{Tanh}   & & \texttt{Sigmoid} & \texttt{ReLU} & \texttt{Tanh}    \\
        \midrule[0.3pt]
        \multirow{4}{*}{$\|\bm{\ell}\|_2^2$}
            & \multirow{2}{*}{Sum}  & -     && 41.67 & 45.03 & 22.02    &&  10.00 & 12.64 & 24.05   & & 10.00 & 12.66 & 41.56    \\
            &                       & WTA  && 41.27 & 45.14 & 33.66    &&  40.26 & 32.69 & 33.50   & & 10.00 & 10.54 & 20.00    \\
            & \multirow{2}{*}{Mean} & -     & &42.46 & 45.86 & 21.49    &&  14.27 & 48.35 & 19.16   & & 33.74 & 42.17 & 10.43    \\
            &                       & WTA  & &39.84 & 45.53 & 42.00    &&  12.76 & 47.35 & 11.73   & & 10.46 & 40.97 & 11.68    \\
            \midrule
        \multirow{4}{*}{$\|\bm{\ell}\|_2$}
            & \multirow{2}{*}{Sum}  & -     & &39.43 & 40.29 & 22.10     & & 14.75 & 18.41 & 34.14   &&  10.00 & 17.20 & 25.51   \\
            &                       & WTA  && 36.17 & 42.52 & 29.14     &&  37.93 & 47.71 & 31.04   &&  12.74 & 16.17 & 28.24   \\
            & \multirow{2}{*}{Mean} & -     & &39.82 & 42.59 & 20.92     & & 10.00 & 49.92 & 13.81   &&  20.84 & 42.81 & 10.43   \\
            &                       & WTA  & &39.23 & 43.89 & 34.30     & & 11.06 & 45.77 & 12.83   &&  10.44 & 39.14 & 12.55   \\
            \midrule
        \multirow{4}{*}{$\|\bm{\ell}\|_1$}
            & \multirow{2}{*}{Sum}  & -     & &42.99 & 46.39 & 12.61     &&  10.00 & 10.03 & 26.72   &&  10.00 & 40.06 & 10.00   \\
            &                       & WTA  && 40.14 & 44.32 & 32.69     &&  40.34 & 23.77 & 38.36   &&  10.00 & 17.03 & 11.51   \\
            & \multirow{2}{*}{Mean} & -     & &43.28 & 45.43 & 12.37     &&  14.21 & 46.51 & 9.38   &&  29.26 & 40.72 & 10.53   \\
            &                       & WTA  & &40.15 & 44.15 & 36.05     &&  11.00 & 34.73 & 14.68   &&  10.10 & 28.46 & 12.33   \\
            
         \midrule[0.3pt]\bottomrule[1pt]
    \end{tabular}
    }
\end{table*}

\section{Additional Results for RQ2: Latent Space Taxonomy}
\label{app:rq2_non_separability}

% A priori bien
This appendix provides an extensive overview of the various latent structures that emerge depending on the neural configurations used to train the models. As presented in the results of RQ1 in Subsection \ref{sec:subsec_rq1}, neurons do not behave uniformly across the different configurations employed during training, which results in different generalization capabilities. This distinct dynamics appear to be closely linked to the geometry manifested in their latent space, resembling the strategies employed by the models to attain their optimal accuracies. The phenomenon of latent spaces acquiring a geometric structure as learning progresses was already observed by Tosato et al. \cite{tosato2023emergent} and by Ororbia \& Mali \cite{ororbia2023predictive}, however, this work extends their analysis to a broader range of activation and probability functions, aiming to showcase the diverse dynamics that FFA-like algorithm can exhibit.

% A priori bien
Given the extensive set of models trained for this study, totaling over 500 models, we conducted a qualitative analysis on their latent space. From this analysis, we selected a small subset of representative spaces that illustrate the features characterizing their respective learning dynamics. Our analysis focused on models achieving more than 50\% accuracy to ensure clarity in the observed geometric patterns. Additionally, for each characteristic latent space described, we provide a brief discussion analyzing the dynamics contributing to the emergence of these latent structures. A detailed depiction of the different latent spaces is provided in the supplementary material.

\begin{figure}[h]
    \centering
    \includegraphics[width=0.9\columnwidth]{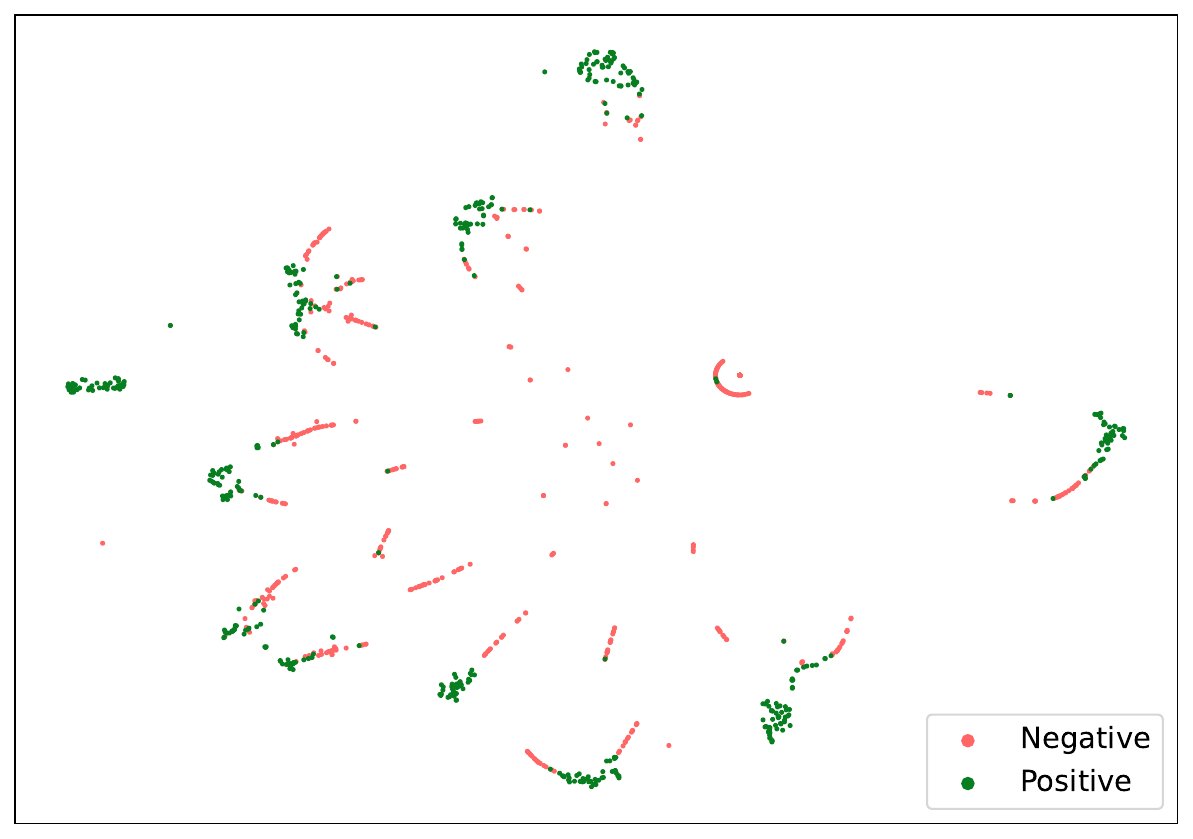}
    \caption{Latent space projection using T-SNE of a network trained with FFA using ReLU activations and a $L_1$ norm with WTA lateral inhibition.}
    \label{fig:app_F_original_ffa}
    
\end{figure}

% A priori bien
Throughout our analysis, we identified five distinct latent structures generated by FFA-like algorithms, each driven by a specific set of neural configurations. The classification of these latent structures is as follows:

% A priori bien
\paragraph{Original FFA} The neural configuration of the original FFA involved using the $L_2$ norm over a network composed of ReLU-activated neurons, integrated with a sigmoidal probability function using a threshold value of $\theta=2$. The resulting geometric structure is characterized by a large cluster of points neighboring zero, representing samples detected as negative, alongside a set of clusters located farther away, each comprising points classified by the model as positive (see Figure \ref{fig:app_F_original_ffa}). Each positive cluster predominantly consists of points belonging to distinct classes, aligning with the effect observed by Tosato et al., where latent vectors of datapoints from the same class converged into similar representations \cite{tosato2023emergent}. This trend does not seem to be exclusive to this configuration, as most latent geometries appear to exhibit this neural specialization phenomenon, indicating that neurons tend to exhibit neural activity primarily when presented with samples from their respective class.

\begin{figure}[h]
    \centering
    \includegraphics[width=0.9\columnwidth]{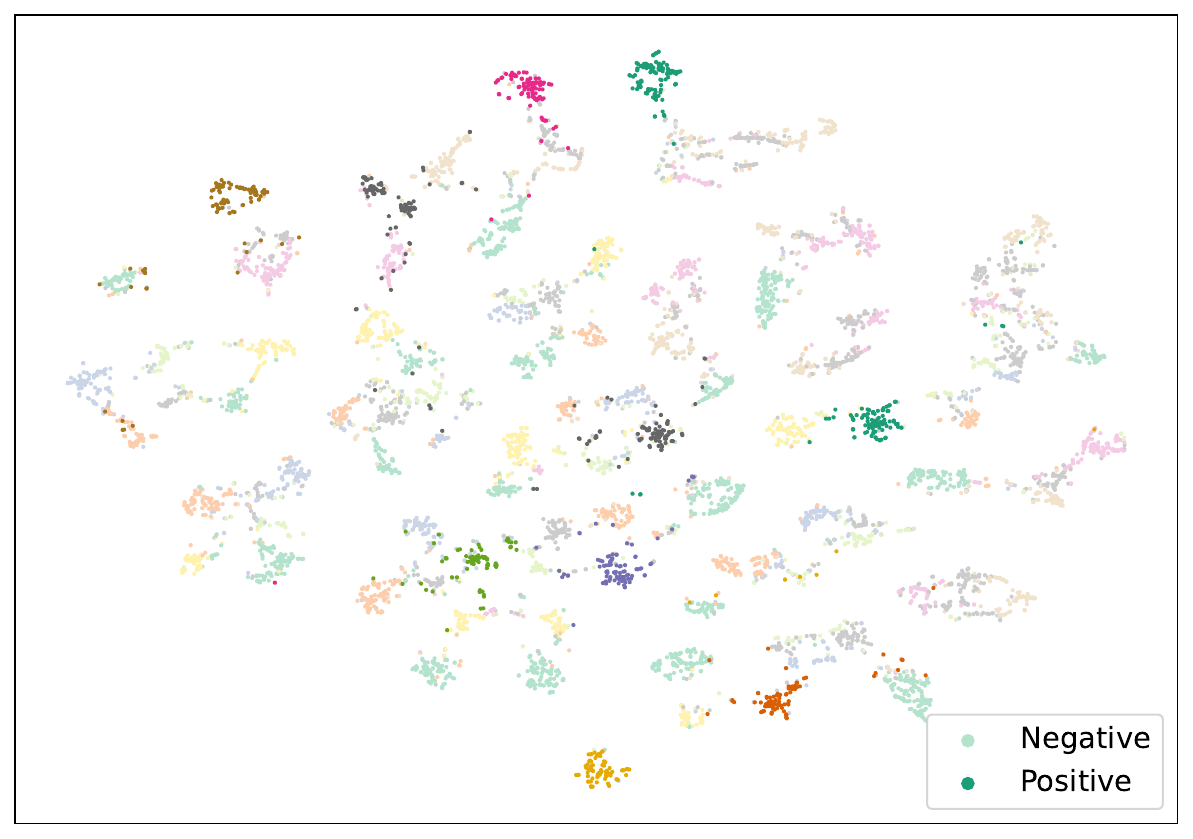}
    \caption{Latent space projection using T-SNE of a network trained with FFA using Sigmoid activations and a $L_1$ norm. Darker colors represent positive samples, while pale colors represent negative inputs.}
    \label{fig:app_F_sigmoid_ffa}
    \vspace{3mm}
\end{figure}

\paragraph{Sigmoid-Activated FFA} In contrast to FFA models trained using ReLU, those employing the sigmoid activation function fail to cluster negative samples within the neighboring area of $0$, as depicted in Figure \ref{fig:app_F_sigmoid_ffa}. This effect arises from the non-negative output range of the sigmoid, where achieving activities close to zero requires extremely low pre-activation values. Consequently, the learning dynamic of this model shifts the objetive from having negative latent vectors close to zero to aiming for positive samples with slightly higher goodness scores than negative ones. As a result, the difference in mean activity between positive and negative vectors becomes evident, while negative vectors still produce latent vectors that closely resemble those generated by positive samples. Usually, the small clusters observed adhere to a correspondence where samples from the same pair of real class and label embedding group together. Unlike ReLU-based FFA networks, latent vectors in this category do not result in sparse vectors. Instead, they maximize their latent activity by activating a large set of neurons close to attaining their maximal value at each coordinate, as dictated by the upper bound of the sigmoid. This lack of sparsity also implies a lesser degree of neural specialization, as many neurons become specialized in a wide range of classes.

\paragraph{Perturbation-Based FFA} Models associated with this latent geometry exhibit less predictable neural configurations. Currently, instances of this structure appear to emerge in cases utilizing the ReLU activation function together with a sum-based aggregation. This structure seems to arise from large values of the ReLU activation resulting in smaller weight changes, primarily focusing on the weights of the label. For example, such spaces can be observed in cases using the $L_2$ norm in MNIST but fail to appear when employing a Winner-Takes-All (WTA) inhibition mechanism, which drastically reduces layer activity. In the latter case, the resulting geometry aligns with what is expected for the original FFA.

\begin{figure}[h]
    \centering
    \includegraphics[width=0.9\columnwidth]{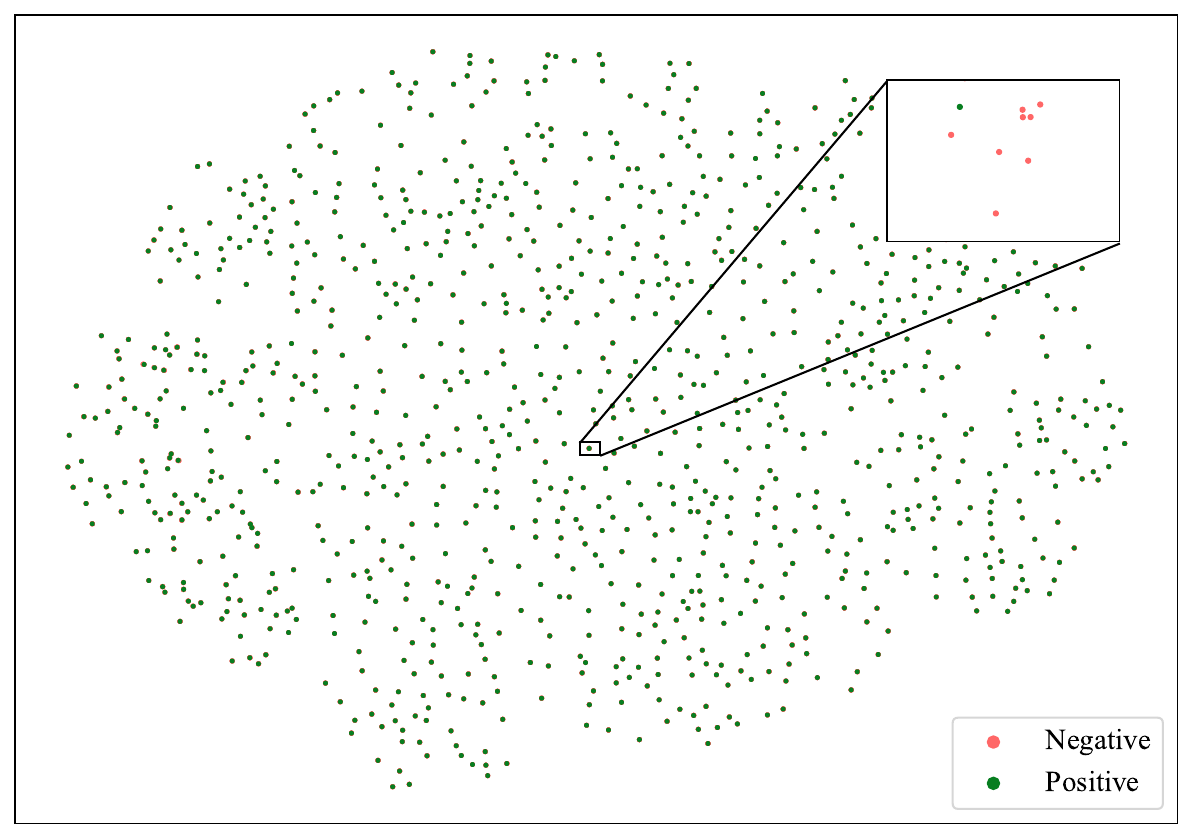}
    \caption{Latent space projection using T-SNE of a network trained with FFA using ReLU activations and a squared $L_2$ norm.}
    \label{fig:app_F_perturbed_ffa}
    
\end{figure}

The geometry arising under this category is characterized by showing no significant separation between positive and negative samples. As depicted in Figure \ref{fig:app_F_perturbed_ffa}, the latent space consists of numerous small clusters, corresponding to the total number of input samples. Additionally, each small cluster contains only $10$ or fewer points, as shown in the small frame in the upper right corner, which presents a zoomed version of one of these clusters. This behavior can be attributed to the reduced impact of the label on the latent vector. In most models, the weights of the latent vector serve to heavily influence the latent in a certain direction. In these cases, the label weights are relatively low, and therefore, the latent is mainly positioned based on the latent generated by the input alone. Formally, we can view this effect as having two independent weight matrices $(\mathbf{W}_{x}, \mathbf{W}_l)$, where $\mathbf{W}_{\mathbf{x}}$ represents the matrix that interacts with the input image $\mathbf{x}$ and $W_l$ interacts with the embedded label $\mathbf{l}$. Since the latent vector is given by $f(\mathbf{W}_{x}\mathbf{x}^T+\mathbf{W}_{l}\mathbf{l}^T)$, if the matrix $\mathbf{W}_l$ has low-valued entries, the value will be approximately given by $f(\mathbf{W}_{x}\mathbf{x}^T+\mathbf{W}_{l}\mathbf{l}^T) \approx f(\mathbf{W}_{x}\mathbf{x}^T)$. However, models using this configuration still achieve good accuracy, as the small perturbation given by the value of $\mathbf{W}_{l}\mathbf{l}^T$ serves to point in the direction that maximizes the latent activity for each class. 

\paragraph{Sigmoid Probability Polar-FFA} Similar to the original FFA, models trained using the sigmoid probability in Polar-FFA generally achieve highly clustered positive latent vectors, which also have a high degree of separation with respect to negative samples. However, due to the additional objective of maximizing the negative neural set when being presented with a negative input, the negative set of latent vectors exhibits positive-like activity. In this case, the information loss created by sending all negatives to zero is replaced by large clouds of latent points, each containing information about the given negative sample. While the cluster structure makes this latent structure partially similar to the one seen in FFA models trained with sigmoid activations, the activity in the negative samples in these models does not overlap with the one present in the positive neuron group, resulting in a clearer separation between positive and negative samples and less overlap.

\begin{figure}[h]
    \centering
    \includegraphics[width=0.9\columnwidth]{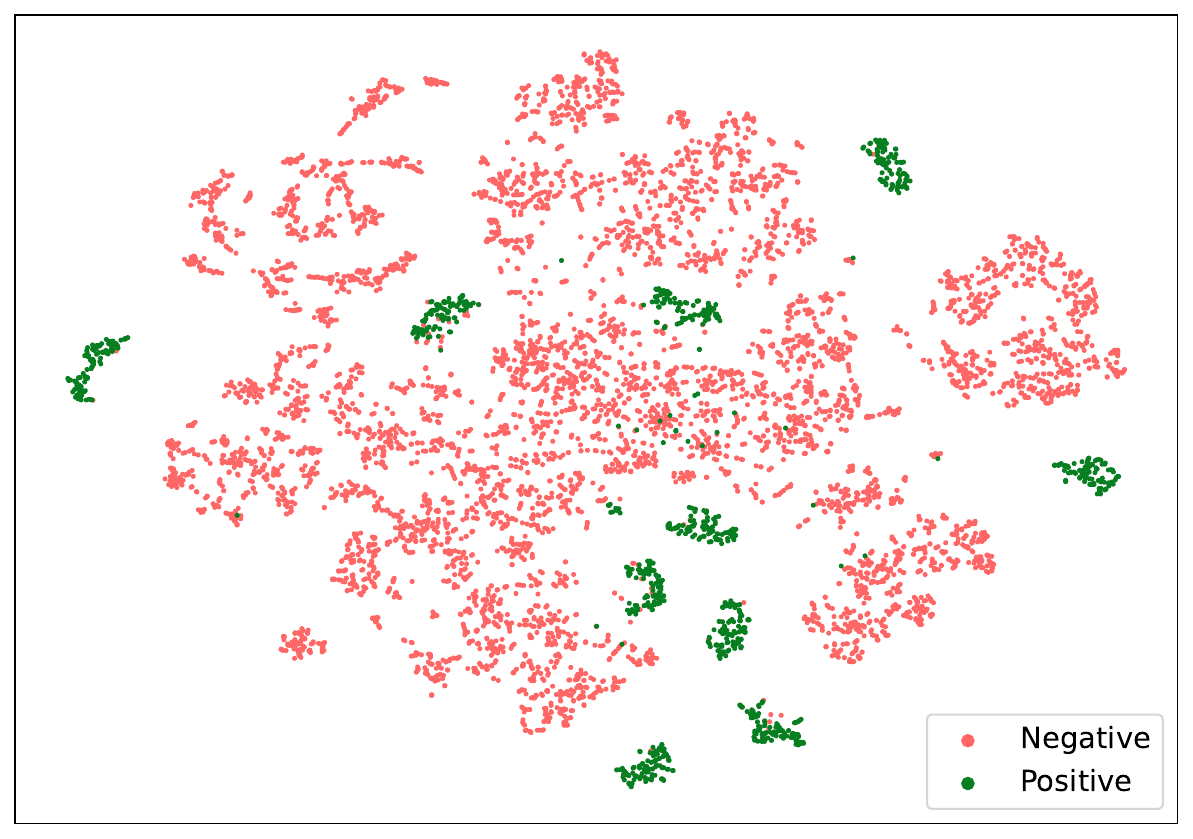}
    \caption{Latent space projection using T-SNE of a network trained with FFA using ReLU activations, the squared $L_2$ norm with WTA lateral inhibition, and the sigmoid probability function.}
    \label{fig:app_F_sigmoid_polar_ffa}
    
\end{figure}

In contrast to the previous cases, where networks exhibited completely different activity when not using ReLU, this effect is not restricted to any specific activation function. For instance, similar behavior can be observed in networks trained using the Tanh activation or the Sigmoid function. These cases result in similar topologies, characterized by clusters based on real class and embedded label, akin to the sigmoid case of FFA. However, Tanh and Sigmoid appear to achieve tighter clustering, possibly due to a reduced variance in the latent vector caused by the bounded behavior of the activation functions.

\begin{figure}[h]
    \centering
    \includegraphics[width=0.9\columnwidth]{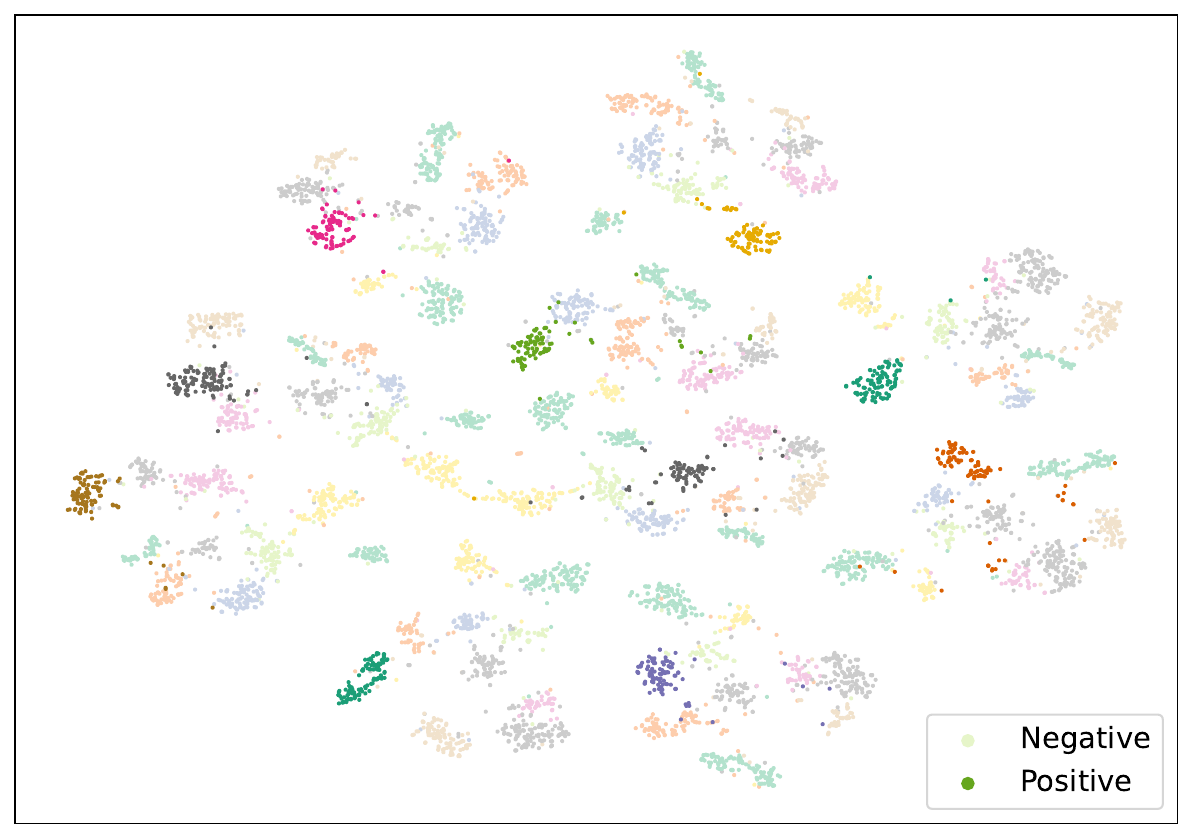}
    \caption{Latent space projection using T-SNE of a network trained with Polar-FFA using Tanh activations, the $L_2$ norm, and the sigmoid probability function. Each color represent the class of the input image. Darker colors represent positive samples, while pale colors represent negative inputs.}
    \label{fig:app_F_sigmoid_polar_ffa_tanh}
    
\end{figure}

\begin{figure}[h]
    \centering
    \includegraphics[width=0.9\columnwidth]{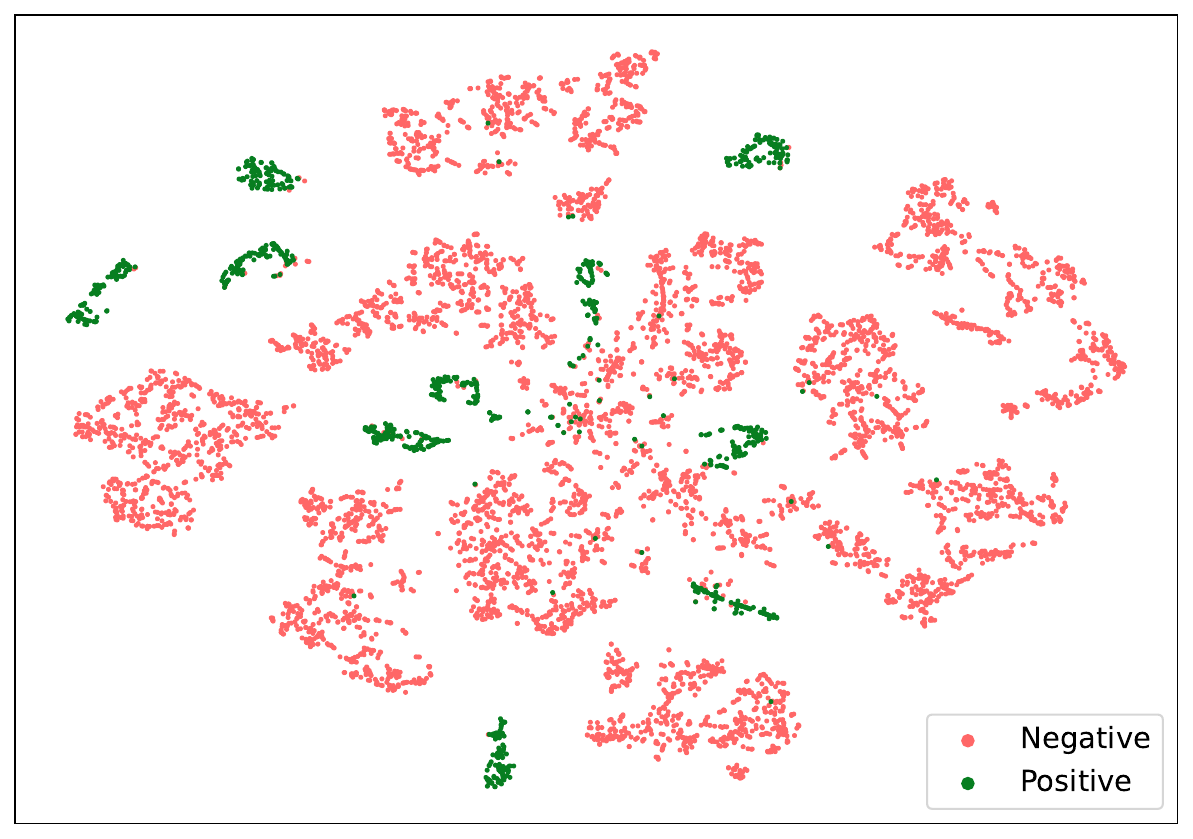}
    \caption{Latent space projection using T-SNE of a network trained with FFA using ReLU activations, the squared $L_2$ norm with WTA lateral inhibition, and the symmetric probability function.}
    \label{fig:app_F_siymm_polar_ffa_tanh}
    
\end{figure}

\begin{figure}[h]
    \centering
    \includegraphics[width=0.9\columnwidth]{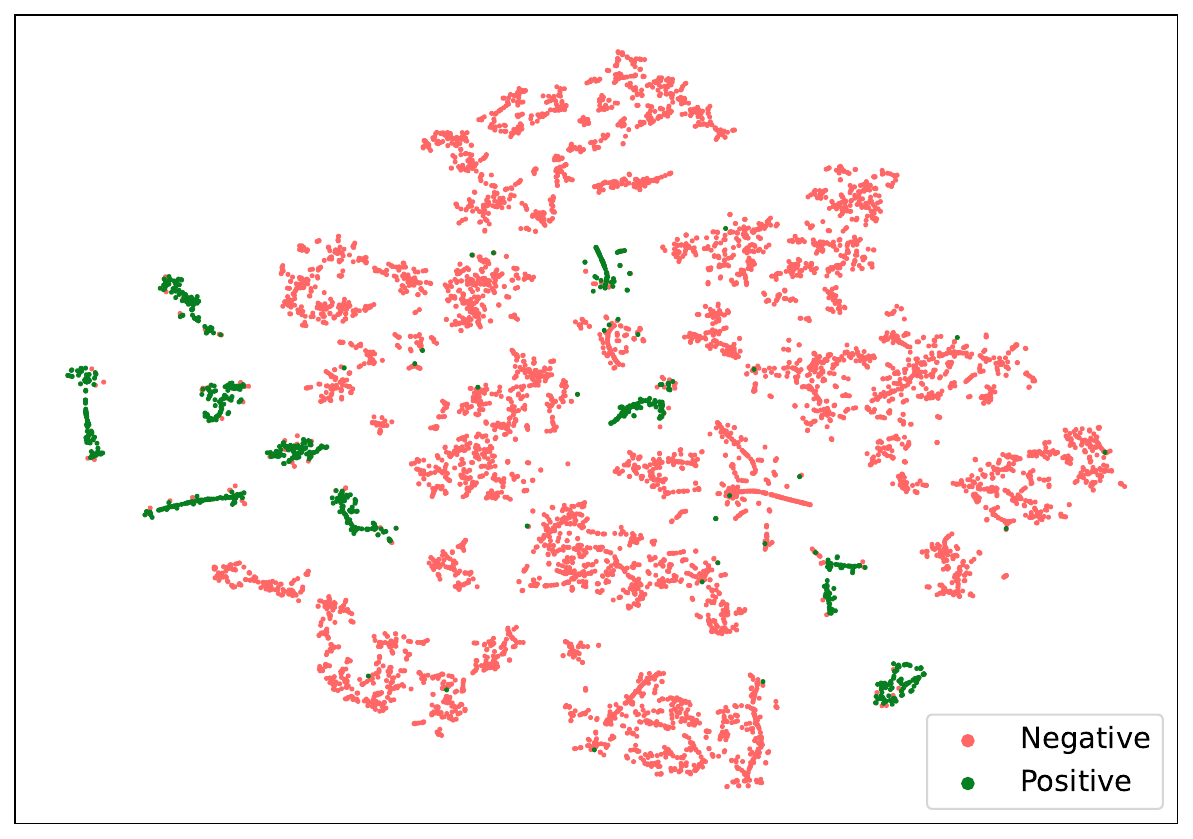}
    \caption{Normalized latent space projection using T-SNE of a network trained with FFA using ReLU activations, the squared $L_2$ norm with WTA lateral inhibition, and the symmetric probability function.}
    \label{fig:app_norm_F_siymm_polar_ffa_tanh}
    
\end{figure}

\paragraph{Symmetric Polar-FFA} Models trained using the symmetric probability function in Polar-FFA appear to generate highly homogeneous latent spaces, as depicted in Figure \ref{fig:app_F_siymm_polar_ffa_tanh}, mostly independent of the activation or the goodness function. This homogeneous latent structure seems to correlate with the previously presented results indicating that these models have the highest generalization capabilities and the most robustness across different seeds. Similar to most other configurations, this probability function generates a high degree of separation between positive and negative samples. However, a noticeable clustering can be observed near the origin, where positive and negative points become clustered together. This clustering arises due to the scale-invariance of these models, which is not accounted for in this depiction. Employing a normalization scheme, as depicted in Figure \ref{fig:app_norm_F_siymm_polar_ffa_tanh}, helps achieve a clearer separation between positive and negative samples. These clusters indicate that each class is related to a direction in the space, which is relative to points in the projective space.

\end{document}